%% file: TPAMI_main_rebuttal.tex
\documentclass[lettersize,journal]{IEEEtran}
\usepackage{amsmath,amsfonts}
\usepackage{algorithmic}
\usepackage{algorithm}
\usepackage{array}
\usepackage[caption=false,font=normalsize,labelfont=sf,textfont=sf]{subfig}
\usepackage{textcomp}
\usepackage{stfloats}
\usepackage{url}
\usepackage{verbatim}
\usepackage{graphicx}
\usepackage{cite}
\usepackage{hyperref}
\usepackage{xcolor}
\usepackage{color}
\usepackage{booktabs}
\usepackage{multirow}
\usepackage{epsfig}
\usepackage{colortbl}
\usepackage{makecell}
\usepackage{caption}
\usepackage{float}
\usepackage{placeins}
\usepackage[marginal]{footmisc}
\usepackage{arydshln} 
\usepackage{caption}
\usepackage{float}

\definecolor{mycolor_blue}{HTML}{E7EFFA}
\definecolor{mycolor_green}{HTML}{E6F8E0}
\definecolor{mycolor_gray}{HTML}{ECECEC}
\definecolor{mycolor_red}{HTML}{FFE6E6}
\definecolor{mycolor_yellow}{HTML}{FFFFCC}
\definecolor{mycolor_purple}{HTML}{E6E6FF}
\newcolumntype{R}{>{\color{red}}l}

\definecolor{citecolor}{HTML}{2980b9}
\definecolor{linkcolor}{HTML}{c0392b}
\definecolor{urlcolor}{HTML}{F08080}
\usepackage{hyperref}
\hypersetup{
    colorlinks=true,
    linkcolor=linkcolor,
    citecolor=citecolor, 
    urlcolor=urlcolor
}

\newcommand{\name}{T2I-CompBench++}
\newcommand{\method}{Generative mOdel finetuning with Reward-driven Sample selection}
\newcommand{\abbr}{GORS}

\renewcommand{\footnoterule}{%
  \kern -3pt                      %
  \hrule width 0.6\columnwidth height 0.5pt %
  \kern 2pt                       %
}

\title{T2I-CompBench++: An Enhanced and Comprehensive Benchmark for Compositional Text-to-image Generation}

\author{
    Kaiyi Huang, %
    Chengqi Duan, %
    Kaiyue Sun,
    Enze Xie,
    Zhenguo Li,
    Xihui Liu
    
    \IEEEcompsocitemizethanks{
        \IEEEcompsocthanksitem K. Huang, K. Sun and X. Liu are with The University of Hong Kong. Email: \{huangky, kaiyue\}@connect.hku.hk, xihuiliu@eee.hku.hk.
        \IEEEcompsocthanksitem C. Duan is with The University of Hong Kong, and Tsinghua University. Email: duancq20@mails.tsinghua.edu.cn.
        \IEEEcompsocthanksitem E. Xie and Z. Li are with the Huawei Noah's Ark Lab. Email:\{xie.enze, zhenguo\}@huawei.com.
        \IEEEcompsocthanksitem 
        Corresponding author: Xihui Liu (xihuiliu@eee.hku.hk). 
    }%
}

\author{
    Kaiyi Huang, %
    Chengqi Duan, %
    Kaiyue Sun,
    Enze Xie,
    Zhenguo Li,
    Xihui Liu
\thanks{

K. Huang, K. Sun and X. Liu are with The University of Hong Kong. Email: \{huangky, kaiyue\}@connect.hku.hk, xihuiliu@eee.hku.hk. 

C. Duan is with The University of Hong Kong, and Tsinghua University. Email: duancq24@connect.hku.hk.

E. Xie and Z. Li are with the Huawei Noah's Ark Lab. Email:\{xie.enze, zhenguo\}@huawei.com.

Corresponding author: Xihui Liu (xihuiliu@eee.hku.hk). 

Manuscript created October, 2020; This work was developed by the IEEE Publication Technology Department. This work is distributed under the \LaTeX \ Project Public License (LPPL) ( http://www.latex-project.org/ ) version 1.3. A copy of the LPPL, version 1.3, is included in the base \LaTeX \ documentation of all distributions of \LaTeX \ released 2003/12/01 or later. The opinions expressed here are entirely that of the author. No warranty is expressed or implied. User assumes all risk.
}
}

\IEEEpubid{0000--0000/00\$00.00~\copyright~2021 IEEE}

\let\oldtwocolumn\twocolumn
\renewcommand\twocolumn[1][]{%
    \oldtwocolumn[{#1}{
    \begin{center}
           \includegraphics[width=\textwidth]{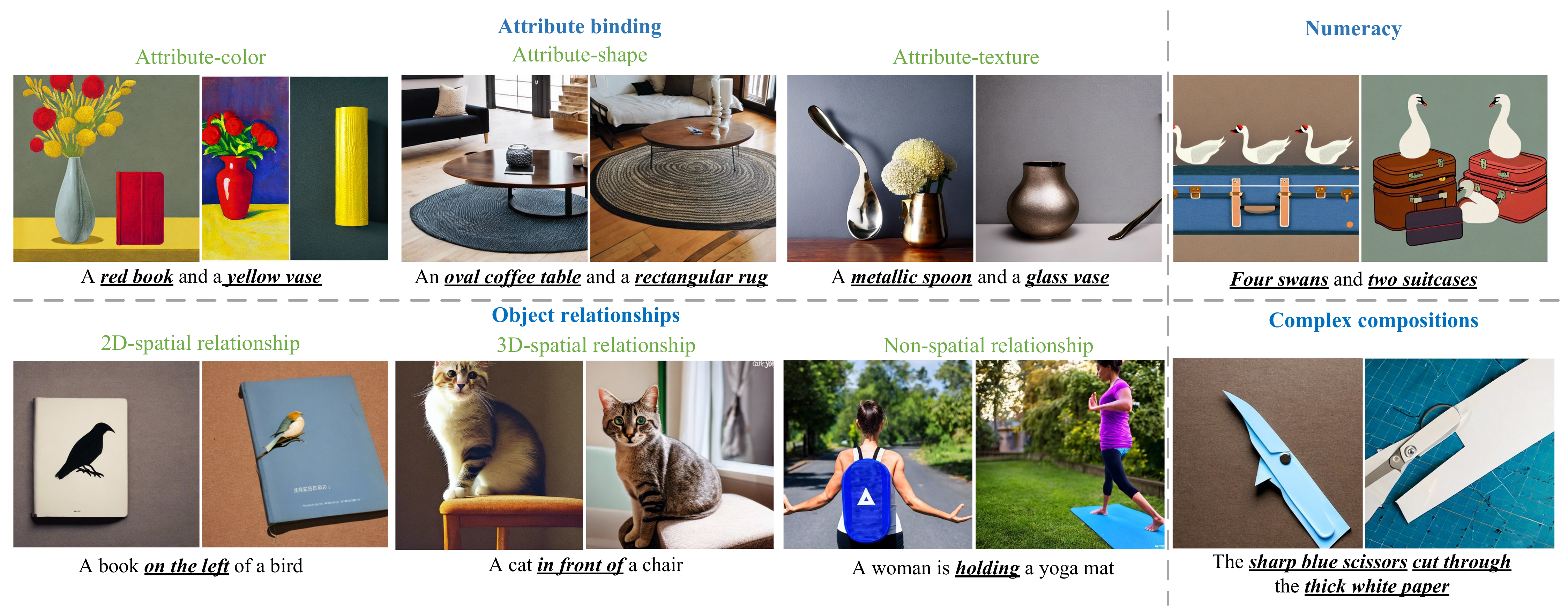}
           \captionof{figure}{
           \textbf{
           Failure cases of Stable Diffusion v2~\cite{rombach2022high}. Our compositional text-to-image generation benchmark consists of three categories: attribute binding (including color, shape, and texture), generative numeracy, object relationships (including 2D/3D-spatial relationship and non-spatial relationship), and complex compositions.}}
           \label{fig:intro}
        \end{center}
    }]
}

\begin{document}
\maketitle

\begin{abstract}

Despite the impressive advances in text-to-image models, they often struggle to effectively compose complex scenes with multiple objects, displaying various attributes and relationships. To address this challenge, we present~\name, an enhanced benchmark for compositional text-to-image generation. \name~comprises 8,000 compositional text prompts categorized into four primary groups: attribute binding, object relationships, generative numeracy, and complex compositions. These are further divided into eight sub-categories, including newly introduced ones like 3D-spatial relationships and numeracy. In addition to the benchmark, we propose enhanced evaluation metrics designed to assess these diverse compositional challenges. These include a detection-based metric tailored for evaluating 3D-spatial relationships and numeracy, and an analysis leveraging Multimodal Large Language Models (MLLMs), \textit{i.e}. GPT-4V, ShareGPT4v as evaluation metrics. Our experiments benchmark 11 text-to-image models, including state-of-the-art models, such as FLUX.1, SD3, DALLE-3, Pixart-$\alpha$, and SD-XL on T2I-CompBench++. We also conduct comprehensive evaluations to validate the effectiveness of our metrics and explore the potential and limitations of MLLMs. Project page is available at \url{https://karine-h.github.io/T2I-CompBench-new/}.

\end{abstract}


\IEEEpubidadjcol
\section{Introduction}
\IEEEPARstart{R}{ecent} progress in text-to-image generation~\cite{ho2022cascaded,rombach2022high,saharia2022photorealistic,dhariwal2021diffusion,nichol2021improved,chang2023muse} has showcased remarkable capabilities in creating diverse and high-fidelity images based on natural language prompts. 
However, we observe that even state-of-the-art text-to-image models often fail to compose multiple objects with different attributes and relationships into a complex and coherent scene, as shown in the failure cases of Stable Diffusion~\cite{rombach2022high} in Figure~\ref{fig:intro}.
For example, given the text prompt ``a blue bench on the left of a green car'', the model might bind attributes to the wrong objects or generate the spatial layout incorrectly.

Previous works have explored compositional text-to-image generation from different perspectives, such as concept conjunction~\cite{liu2022compositional}, attribute binding (focusing on color)~\cite{feng2022training,chefer2023attend}, and spatial relationship~\cite{wu2023harnessing}.
Most of those works focus on a sub-problem and propose their own benchmarks for evaluating their methods. %
However, there is no consensus on the problem definition and standard benchmark of compositional text-to-image generation. Another challenge is the assessment of compositional text-to-image models.
Most previous works evaluate the models by image-text similarity or text-text similarity (between the caption predicted from the generated images and the original text prompts) with CLIPScore~\cite{radford2021learning,hessel2021clipscore} or BLIP~\cite{li2022blip,li2023blip}.
However, both metrics do not perform well for compositionality evaluation due to the ambiguity and difficulty in compositional vision-language understanding. %

In this paper, we propose an enhanced benchmark for compositional text-to-image generation, namely~\name, which is the first comprehensive benchmark that fills a critical gap in evaluating the compositional capabilities of text-to-image generation models. \name~not only provides a robust evaluation framework for assessing compositionality but also drives advancements in generating complex, high-fidelity images from intricate compositional text descriptions.

First, we propose a compositional text-to-image generation benchmark,~\name, which consists of four categories and eight sub-categories of compositional text prompts:
(1) Attribute binding. Each text prompt in this category contains at least two objects and two attributes, and the model should bind the attributes with the correct objects to generate the complex scene. This category is divided into three sub-categories (color, shape, and texture) based on the attribute type.
(2) Object relationships. The text prompts in this category each contain at least two objects with specified relationships between the objects. Based on the type of the relationships, this category consists of three sub-categories, 2D/3D-spatial relationship and non-spatial relationship.
(3) Generative numeracy. Each prompt in this category involves one or multiple object categories with numerical quantities, ranging from one to eight.
(4) Complex compositions, where the text prompts contain more than two objects or more than two sub-categories mentioned above. For example, a text prompt that describes three objects with their attributes and relationships.

Second, we introduce a set of evaluation metrics tailored to different categories of compositional prompts. For attribute binding evaluation, we propose the \textit{Disentangled BLIP-VQA} metric, designed to address the challenges of ambiguous attribute-object correspondences. For assessing spatial relationships and numeracy, we introduce the \textit{UniDet-based metric}, which also incorporates depth estimation techniques and object detection mechanisms to evaluate 3D spatial relationships. Additionally, we propose an \textit{MLLM (Multimodal Large Language Model)-based metric} for non-spatial relationships and complex compositions, examining the performance and limitations of MLLMs such as \textit{ShareGPT4V}~\cite{chen2023sharegpt4v}, and \textit{GPT-4V}~\cite{yang2023dawn} for compositionality evaluation.

Finally, we propose a new approach, \textit{\method~(\abbr)}, for compositional text-to-image generation. We finetune Stable Diffusion v2~\cite{rombach2022high} model with generated images that highly align with the compositional prompts, where the fine-tuning loss is weighted by the reward which is defined as the alignment score between compositional prompts and generated images. This approach is simple but effective in boosting the model's compositional abilities and can serve as a new baseline for future explorations.

Experimental results across four categories and eight subcategories, validated with human correlation, demonstrate the effectiveness of our proposed evaluation metrics. Our method, \abbr, consistently outperforms the baseline models. We benchmark 11 models on~\name, including FLUX.1~\cite{FLUX}, SD3~\cite{esser2024scaling}, DALLE-3~\cite{betker2023improving}, Pixart-$\alpha$~\cite{chen2024pixart}, and SD-XL~\cite{podell2023sdxl}, highlighting the performances of current models in handling compositional tasks.

Compared with the preliminary conference version~\cite{huang2024t2i}, this work introduces several non-trivial extensions in the following aspects:

\begin{itemize}
    \item \textbf{Broadening the problem definition and categories of compositional text-to-image generation.} As the rapid evolvement of text-to-image generation models in the past year call for more comprehensive benchmarks, we broaden the problem definition of our benchmark by adding two sub-categories: generative numeracy (e.g., ``four swans and two suitcases'') and 3D-spatial relationships (e.g.,``a cat in front of a chair''), resulting in a more enhanced benchmark with four categories and eight sub-categories.
    \item \textbf{Introducing more evaluation metrics and in-depth discussions on MLLMs as evaluation metrics.} We introduce evaluation metrics specifically tailored for the newly added categories of numeracy and 3D-spatial relationships. Utilizing depth estimation techniques and object detection mechanisms, these metrics offer a robust framework for assessing performance across various compositional domains. Additionally, we analyzed the capabilities of advanced Multimodal Large Language Models (MLLMs), such as ShareGPT4V and GPT-4V, focusing on their effectiveness in addressing compositional challenges. We assess MLLM through comparing human correlation across eight sub-categories and examining their performances of stability over multiple executions.
    \item \textbf{More comprehensive benchmarks and analysis.} As the community of text-to-image generation has been evolving rapidly and many new foundation text-to-image generation models have emerged in the past year, apart from the SD v2~\cite{rombach2022high} models benchmarked in the conference version, we conduct extensive benchmarks on 11 text-to-image models, including state-of-the-art examples like FLUX.1~\cite{FLUX}, SD3~\cite{esser2024scaling}, DALLE-3~\cite{betker2023improving}, Pixart-$\alpha$~\cite{chen2024pixart}, and SD-XL~\cite{podell2023sdxl}. Our analysis provides deeper insights into current performances and limitations of both T2I models. We believe such extensive benchmarks and analysis will shed light on future works on improving compositionality of text-to-image generation models and inspire future research in visual content generation. 
\end{itemize}

\input{sections/related}

\input{sections/dataset}

\input{sections/evaluation}

\input{sections/method}

\input{sections/experiments}

\bibliographystyle{IEEEtran}
\bibliography{ref}

\clearpage
\input{sections/appendix_rebuttal}

\end{document}

%% file: sections/related.tex
\section{Related work}
\label{sec:related}

\begin{table*}[t]
\caption{Comparison with previous compositional text-to-image benchmarks.}
\label{dataset}
\resizebox{\linewidth}{!}{
\begin{tabular}{lll}
\toprule
\textbf{Benchmark} & \textbf{Prompts number and coverage} & \textbf{Vocabulary diversity} \\ \midrule
CC-500~\cite{feng2022training} & 500 attr bind (color) & 196 nouns, 12 colors \\
ABC-6K~\cite{feng2022training} & 6,000 attr bind (color) & 3,690 nouns, 33 colors \\
Attn-Exct~\cite{chefer2023attend} & 210 attr bind (color) & 24 nouns, 11 colors \\
HRS-comp~\cite{bakr2023hrs} & 1,000 attr bind (color, size), 2,000 rel (spatial, action) & 620 nouns, 5 colors, 11 spatial, 569 actions \\ \midrule
\name & \begin{tabular}[l]{@{}l@{}}3,000 attr bind (color, shape, texture)\\ 3,000 rel (2D/3D spatial, non-spatial) \\
1,000 numeracy\\ 1,000 complex\end{tabular} & \begin{tabular}[l]{@{}l@{}}2,470 nouns, 33 colors, 32 shapes, 23 textures\\ 7 2D-spatial rel, 3 3D-spatial rel, 875 non-spatial rel\end{tabular} \\ \bottomrule
\end{tabular}
}
\end{table*}

\textbf{Text-to-image generation.}
Early works~\cite{reed2016generative,reed2016learning,zhang2017stackgan,xu2018attngan,zhu2019dm,zhang2021cross} explore different network architectures and loss functions based on generative adversarial networks (GAN)~\cite{goodfellow2014generative}.
Recently, diffusion models have achieved remarkable success for text-to-image generation~\cite{ramesh2022hierarchical,nichol2022glide,rombach2022high,saharia2022imagen,gafni2022make, chen2024pixart, gu2023matryoshka,ramesh2021zero, betker2023improving}.
By training on web-scale data, T2I models such as Stable Diffusion~\cite{rombach2022high, podell2023sdxl, esser2024scaling}, DALL-E 3~\cite{betker2023improving}, MDM~\cite{gu2023matryoshka}, and Pixart-$\alpha$~\cite{chen2024pixart}, have
shown remarkable generative power. 
Current state-of-the-art models such as Stable Diffusion~\cite{rombach2022high} still struggle to compose multiple objects with attributes and relationships in a complex scene.
Some recent works attempt to align text-to-image models with human feedback~\cite{zhang2023hive,lee2023aligning}. RAFT~\cite{dong2023raft} proposes reward-ranked fine-tuning to align text-to-image models with certain metrics.
Our proposed \abbr~approach is a simpler finetuning approach that does not require multiple iterations of sample generation and selection.

\textbf{Compositional text-to-image generation.}
Researchers have delved into various aspects of compositionality in text-to-image generation to achieve visually coherent and semantically consistent results~\cite{liu2022compositional,feng2022training,li2022stylet2i,wu2023harnessing, huang2024t2i, patel2023eclipse, liu2024referee}. 
Previous work focused on concept conjunction and negation~\cite{liu2022compositional}, attribute binding with colors~\cite{feng2022training,chefer2023attend,park2021benchmark}, generative numeracy~\cite{lian2023llm}, and spatial relationships between objects~\cite{chen2024training,wu2023harnessing}.
However, those work each target at a sub-problem, and evaluations are conducted in constrained scenarios. 
Recent compositional studies typically fall into two categories~\cite{wang2023compositional}: one relies on cross attention maps for compositional generation~\cite{meral2023conform, kim2023dense, rassin2024linguistic}, while the other integrates layout as a generation condition~\cite{gani2023llm, li2023gligen, taghipour2024box, wang2024divide, chen2023reason}. LMD~\cite{lian2023llm}, RPG~\cite{yang2024mastering}, and RealCompo~\cite{zhang2024realcompo} utilize LLMs or MLLMs to reason out layouts or decompose compositional problems.
Our work is the first to introduce a comprehensive benchmark for compositional text-to-image generation.

\textbf{Benchmarks for text-to-image generation.}
Early works evaluate text-to-image on CUB birds~\cite{wah2011caltech}, Oxford flowers~\cite{nilsback2008automated}, and COCO~\cite{lin2014microsoft} which are easy with limited diversity.
As the text-to-image models become stronger, more challenging benchmarks have been introduced.
DrawBench~\cite{saharia2022photorealistic} consists of 200 prompts to evaluate counting, compositions, conflicting, and writing skills. %
DALL-EVAL~\cite{cho2023dall} proposes PaintSkills to evaluate visual reasoning skills, image-text alignment, image quality, and social bias by 7,330 prompts.
HE-T2I~\cite{petsiuk2022human} proposes 900 prompts to evaluate counting, shapes, and faces for text-to-image. %
Several compositional text-to-image benchmarks have also been proposed.
Park \textit{et al.}~\cite{park2021benchmark} proposes a benchmark on CUB Birds~\cite{wah2011caltech} and Oxford Flowers~\cite{nilsback2008automated} to evaluate the models' ability to generate images with object-color and object-shape compositions. %
ABC-6K and CC500~\cite{feng2022training} benchmarks are proposed to evaluate attribute binding for text-to-image models, but they only focused on color attributes.
HRS-Bench~\cite{bakr2023hrs} is a general-purpose benchmark that evaluates 13 skills with 45,000 prompts. Compositionality is only one of the 13 evaluated skills which is not extensively studied.
We propose the first comprehensive benchmark for open-world compositional text-to-image generation, shown in Table~\ref{dataset}. %

\textbf{Evaluation metrics for text-to-image generation.}
Existing metrics for text-to-image generation can be categorized into fidelity assessment, alignment assessment, and LLM-based metrics. 
Traditional metrics such as Inception Score (IS)~\cite{salimans2016improved} and Frechet Inception Distance (FID)~\cite{heusel2017gans} are commonly used to evaluate the fidelity of synthesized images.
To assess the image-text alignment, text-image matching by CLIP~\cite{radford2021learning} and BLIP2~\cite{li2023blip} and text-text similarity by BLIP~\cite{li2022blip} captioning and CLIP text similarity are commonly used. %
Some works leverage the strong reasoning abilities of large language models (LLMs) for evaluation~\cite{lu2023llmscore,chen2023xiqe, wen2023improving}.
Besides, human preferences or feedbacks are also included in text-to-image generation evaluation~\cite{xu2023imagereward, sun2023dreamsync, kirstain2024pick, wu2023better, wu2023human, liang2023rich, ku2023imagenhub, ku2023viescore}. More fine-grained metrics are proposed in~\cite{lee2024holistic}.
However, there was no comprehensive study on how well those evaluation metrics work for compositional text-to-image generation. We propose evaluation metrics specifically designed for our benchmark and validate that our proposed metrics align better with human perceptions.%

%% file: sections/dataset.tex
\section{\name}
\label{headings}

Compositionality of text-to-image models refers to the ability of models to compose different concepts into a complex and coherent scene according to text prompts. %
To provide a clear definition of the problem and to build our benchmark, we introduce four categories and eight sub-categories of compositionality, attribute binding (including three sub-categories: color, shape, and texture), object relationships (including three sub-categories: 2D/3D-spatial relationship and non-spatial relationship), numeracy, and complex compositions.
We generate 1,000 text prompts (700 for training and 300 for testing) for each sub-category, resulting in 8,000 compositional text prompts in total.
We take the balance between seen \textit{v.s.} unseen compositions in the test set, prompts with fixed sentence template \textit{v.s.} natural prompts, and simple \textit{v.s.} complex prompts into consideration when constructing the benchmark.
The text prompts are generated with either predefined rules or ChatGPT~\cite{ChatGPT}, so it is easy to scale up.

It is important to note that prompts generated by ChatGPT~\cite{ChatGPT} may include scenarios that do not conform to physical laws. This is an intentional design choice in our dataset to challenge models with both common and uncommon compositions, including those that break real-world constraints. This consideration is essential for evaluating and advancing the compositional capabilities of text-to-image generative models, ensuring they do not just memorize the dataset but can generalize to unseen combinations.

\begin{figure}[t]
    \centering
    \includegraphics[width=\linewidth]{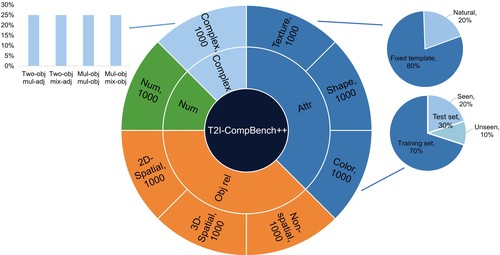}
    \caption{Data statistics of~\name.}
    \label{fig:data}
\end{figure}

Comparisons between our benchmark and previous benchmarks are shown in Table.~\ref{dataset}, and data statistics in Figure~\ref{fig:data}.

\subsection{Attribute Binding}
A critical challenge for compositional text-to-image generation is attribute binding, where attributes must be associated with corresponding objects in the generated images. 
We find that models tend to confuse the association between attributes and objects when there are more than one attribute and more than one object in the text prompt.
For example, with the text prompt ``A room with blue curtains and a yellow chair'', the text-to-image model might generate a room with yellow curtains and a blue chair. 
We introduce three sub-categories, color, shape, and texture, according to the attribute type, and construct 1000 text prompts for each sub-category.
For each sub-category, there are 800 prompts with the fixed sentence template ``{a} \{adj\} \{noun\} and {a} \{adj\} \{noun\}'' (\textit{e.g.}, ``a red flower and a yellow vase'') and 200 natural prompts without predefined sentence template (\textit{e.g.}, ``a room with blue curtains and a yellow chair''). The 300-prompt test set of each sub-category consists of 200 prompts with seen adj-noun compositions (adj-noun compositions appeared in the training set) and 100 prompts with unseen adj-noun compositions (adj-noun compositions not in the training set).

\textbf{Color.}
Color is the most commonly-used attribute for describing objects in images, and current text-to-image models often confuse the colors of different objects.
The 1,000 text prompts related to color binding are constructed with 480 prompts from CC500~\cite{feng2022training}, 200 prompts from COCO~\cite{lin2014microsoft}, and 320 prompts generated by ChatGPT.

The prompt for ChatGPT is: \textit{Please generate prompts in the format of ``{a} \{adj\} \{noun\} and {a} \{adj\} \{noun\}'' by using the color adj. , such as ``a green bench and a red car''.}

\textbf{Shape.}
We define a set of shapes that are commonly used for describing objects in images: long, tall, short, big, small, cubic, cylindrical, pyramidal, round, circular, oval, oblong, spherical, triangular, square, rectangular, conical, pentagonal, teardrop, crescent, and diamond.
We provide those shape attributes to ChatGPT and ask ChatGPT to generate prompts by composing those attributes with arbitrary objects, for example, ``a rectangular clock and a long bench''.

For fixed sentence template, the prompt for ChatGPT is: \textit{Please generate prompts in the format of ``{a} \{adj\} \{noun\} and {a} \{adj\} \{noun\}'' by using the shape adj.: long, tall, short, big, small, cubic, cylindrical, pyramidal, round, circular, oval, oblong, spherical, triangular, square, rectangular, conical, pentagonal, teardrop, crescent, and diamond.}
For natural prompts, the prompt for ChatGPT is: \textit{Please generate objects with shape adj. in a natural format by using the shape adj.: long, tall, short, big, small, cubic, cylindrical, pyramidal, round, circular, oval, oblong, spherical, triangular, square, rectangular, conical, pentagonal, teardrop, crescent, and diamond.}

\textbf{Texture.}
Textures are also commonly used to describe the appearance of objects. They can capture the visual properties of objects, such as smoothness, roughness, and granularity.
We often use materials to describe the texture, such as wooden, plastic, and rubber.
We define several texture attributes and the objects that can be described by each attribute.
We generate 800 text prompts by randomly selecting from the possible combinations of two objects each associated with a textural attribute, \textit{e.g.}, ``A rubber ball and a plastic bottle''.
We also generate 200 natural text prompts by ChatGPT.

We generate 200 natural text prompts by ChatGPT with the following prompt: \textit{Please generate objects with texture adj. in a natural format by using the texture adj.: rubber, plastic, metallic, wooden, fabric, fluffy, leather, glass.}
Besides the ChatGPT-generated text prompts, we also provide the predefined texture attributes and objects that can be described by each texture, as shown in Table~\ref{texture_dataset}. 
We generate 800 text prompts by randomly selecting from the possible combinations of two objects each associated with a textural attribute, \textit{e.g.}, ``A rubber ball and a plastic bottle''.

\input{tables/tab_texture_dataset}

\subsection{Object Relationship}
When composing objects in a complex scene, the relationship between objects is a critical factor. %
We introduce 1,000 text prompts for 2D/3D-spatial relationships and non-spatial relationships, respectively.

\textbf{2D-spatial relationships.}
We use ``on the side of'', ``next to'', ``near'', ``on the left of'', ``on the right of'', ``on the bottom of'', and ``on the top of'' to define 2D-spatial relationships. 
The two nouns are randomly selected from persons (\textit{e.g.}, man, woman, girl, boy, person, \textit{etc.}),  animals (\textit{e.g.}, cat, dog, horse, rabbit, frog, turtle, giraffe, \textit{etc.}), 
and objects (\textit{e.g.}, table, chair, car, bowl, bag, cup, computer, \textit{etc.}).
For spatial relationships including left, right, bottom, and top, we construct contrastive prompts by swapping the two nouns, for example, ``a girl on the left of a horse'' and ``a horse on the left of a girl''.

\textbf{3D-spatial relationships.} To illustrate the 3D-spatial relationships, we use “in front of”, “behind” and “hidden by” to define 3D-spatial relationships. Similar to 2D-spatial relationships, we use the same vocabulary to construct the pairs in a prompt, such as “a girl in front of a horse”.

\textbf{Non-spatial relationships.}
Non-spatial relationships usually describe the interactions between two objects. %
We prompt ChatGPT to generate text prompts with non-spatial relationships (\textit{e.g.}, ``watch'', ``speak to'', ``wear'', ``hold'', ``have'', ``look at'', ``talk to'', ``play with'', ``walk with'', ``stand on'', ``sit on'', \textit{etc.}) and arbitrary nouns.%

The prompt for ChatGPT is: \textit{Please generate natural prompts that contain subjects and objects by using relationship words such as wear, watch, speak, hold, have, run, look at, talk to, jump, play, walk with, stand on, and sit on.}

\subsection{Numeracy.}
In the construction of numeracy dataset, we introduce 1,000 prompts, and adopt a tripartite structure based on the quantities of objects. The initial 30\% part of the numeracy contains scenarios involving a singular object, followed by the subsequent 30\% comprising two objects, and the remaining 40\% encompassing multiple objects, with the quantities ranging from one to eight. We construct a list of objects, and randomly combine objects and quantities in a standardized format: ``{number} {object}'' (\textit{e.g.,} ``one pear and three knives''). In order to incorporate diversity of expressions, each part is characterized by a combination of fixed templates and flexible prompts at a ratio of 4:1. 
The Fixed templates adhere to the predefined format, while flexible prompts incorporate natural language expressions encountered in our daily lives with multiple '{number} {object}' included (\textit{e.g.,} ``three plates and three pens were on the table''). 

We construct a list of objects by asking ChatGPT: \textit{Please generate a list of 150 types of objects and their plurals.} Following that, we manually filtered out repeated entries and uncommon objects. For generating flexible prompts, we ask ChatGPT, \textit{Convert to natural prompt: You are given {number} sentences, each containing multiple objects. For each sentence, use the objects in it to create a natural compositional sentence.}

\subsection{Complex Compositions}
To test text-to-image generation approaches with more natural and challenging compositional prompts in the open world, we introduce 1,000 text prompts with complex compositions of concepts beyond the pre-defined patterns. 
Regarding the number of objects, we create text prompts with more than two objects, for example, ``a room with a blue chair, a black table, and yellow curtains''.
In terms of the attributes associated with objects, we can use multiple attributes to describe an object (denoted as \textit{multiple attributes}, \textit{e.g.}, ``a big, green apple and a tall, wooden table''), or leverage different types of attributes in a text prompt (denoted as \textit{mixed attributes}, \textit{e.g.}, the prompt ``a tall tree and a red car'' includes both shape and color attributes).
We generate 250 text prompts with ChatGPT for each of the four scenarios: \textit{two objects with multiple attributes, two objects with mixed attributes, more than two objects with multiple attributes, and more than two objects with mixed attributes}. Relationship words can be adopted in each scenario to describe the relationships among two or more objects.
For each scenario, we split 175 prompts for the training set and 75 prompts for the test set.

(1) For 2 objects with mixed attributes, the prompt for ChatGPT is: 
\textit{Please generate natural compositional phrases, containing 2 objects with each object one adj. from \{color, shape, texture\} descriptions and spatial (left/right/top/bottom/next to/near/on side of) or non-spatial relationships.}
(2) For 2 objects with multiple attributes, the prompt for ChatGPT is: 
\textit{Please generate natural compositional phrases, containing 2 objects with several adj. from \{color, shape, texture\} descriptions and spatial (left/right/top/bottom/next to/near/on side of) or non-spatial relationships.}
(3) For multiple objects with mixed attributes, the prompt for ChatGPT is: 
\textit{Please generate natural compositional phrases, containing multiple objects (number\textgreater 2) with each one adj. from \{color, shape, texture\} descriptions and spatial (left/right/top/bottom/next to/near/on side of) non-spatial relationships.}
(4) For multiple objects with multiple attributes, the prompt for ChatGPT is:  
\textit{Please generate natural compositional phrases, containing multiple objects (number\textgreater 2) with several adj. from \{color, shape, texture\} descriptions and spatial (left/right/top/bottom/next to/near/on side of) or non-spatial relationships.}

%% file: tables/tab_texture_dataset.tex
\begin{table*}[h]
 \centering
 \caption{Textural attributes and associated objects to construct the attribute-texture prompts.}
 \label{texture_dataset}
 \resizebox{\textwidth}{!}{
 \begin{tabular}{lll}
   \toprule
   \cmidrule(r){1-2}
   Textures     & Objects     \\
   \midrule
   Rubber & band, ball, tire, gloves, sole shoes, eraser, boots, mat     \\
   Plastic     & Bottle, bag, toy, cutlery, chair, phone case, container, cup, plate       \\
   Metallic     & car, jewelry, watch, keychain, desk lamp, door knob, spoon, fork, knife, key, ring, necklace, bracelet, earring         \\
   Wooden  & chair, table, picture frame, toy, jewelry box, door, floor, chopsticks, pencils, spoon, knife \\
   Fabric & bag, pillow, curtain, shirt, pants, dress, blanket, towel, rug, hat, scarf, sweater, jacket\\
   Fluffy &pillow, blanket, teddy bear, rug, sweater, clouds, towel, scarf, hat\\
   Leather &jacket, shoes, belt, bag, wallet, gloves, chair, sofa, hat, watch\\
   Glass &bottle, vase, window, cup, mirror, jar, table, bowl, plate\\
   \bottomrule
 \end{tabular}
 }
\end{table*}

%% file: sections/evaluation.tex
\section{Evaluation Metrics}
\label{headings}
\begin{figure*}[t]
    \centering
    \includegraphics[width=\linewidth]{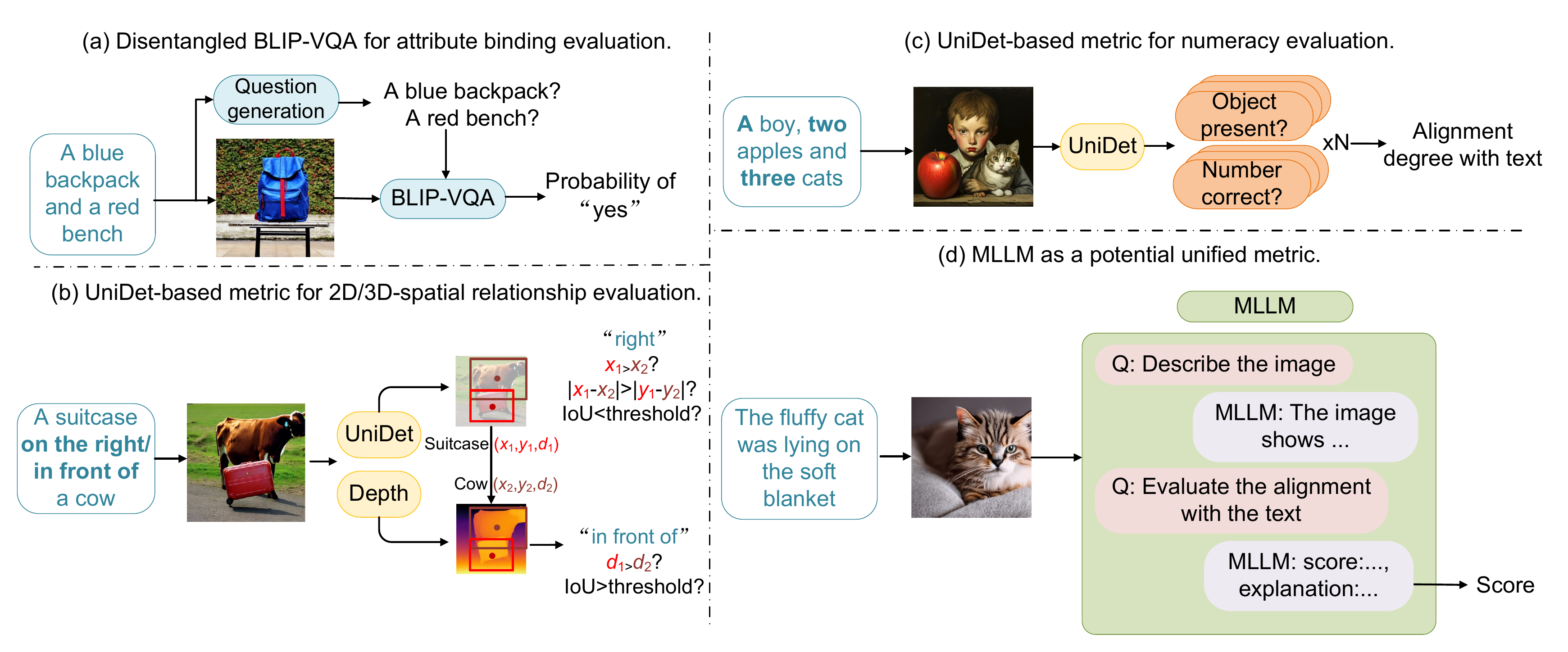}
    \caption{Illustration of our proposed evaluation metrics: (a) Disentangled BLIP-VQA for attribute binding evaluation, (b) UniDet for 2D/3D-spatial relationship evaluation, (c) UniDet for numeracy evaluation, and (d) MLLM as a potential unified metric.}
    \label{fig:pipeline}
\end{figure*}

Evaluating compositional text-to-image generation is challenging as it requires comprehensive and fine-grained cross-modal understanding.
Existing evaluation metrics leverage vision-language models trained on large-scale data for evaluation. 
CLIPScore~\cite{radford2021learning,hessel2021clipscore} calculates the cosine similarity between text features and generated-image features extracted by CLIP.
Text-text similarity by BLIP-CLIP~\cite{chefer2023attend}
applies BLIP~\cite{li2022blip} to generate captions for the generated images, and then calculates the CLIP text-text cosine similarity between the generated captions and text prompts.
Those evaluation metrics can measure the coarse text-image similarity, but fails to capture fine-grained text-image correspondences in attribute binding and spatial relationships. To address those limitations, we propose new evaluation metrics for compositional text-to-image generation, shown in Fig.~\ref{fig:pipeline}. Concretely, we propose \textit{disentangled BLIP-VQA} for attribute binding evaluation, \textit{UniDet-based metric} for 2D/3D-spatial relationship and numeracy evaluation, and 
\textit{MLLM-based metric} for non-spatial relationship and complex prompts.
We further investigate the potential and limitations of multimodal
large language models such as \textit{MiniGPT-4}~\cite{zhu2023minigpt4} with Chain-of-Thought~\cite{wei2022chain}, ShareGPT4V~\cite{chen2023sharegpt4v}, and GPT-4V~\cite{yang2023dawn} for compositionality
evaluation.

\subsection{Disentangled BLIP-VQA for Attribute Binding Evaluation}
We observe that the major limitation of the BLIP-CLIP evaluation is that the BLIP captioning models do not always describe the detailed attributes of each object. For example, the BLIP captioning model might describe an image as ``A room with a table, a chair, and curtains'', while the text prompt for generating this image is ``A room with yellow curtains and a blue chair''. So explicitly comparing the text-text similarity might cause ambiguity and confusion.

Therefore, we leverage the visual question answering (VQA) ability of BLIP~\cite{li2022blip} for evaluating attribute binding.
For instance, given the image generated with the text prompt ``a green bench and a red car'', we ask two questions separately: ``a green bench?'', and ``a red car?''.
By explicitly disentangling the complex text prompt into two independent questions where each question contains only one object-attribute pair, we avoid confusion of BLIP-VQA.
The BLIP-VQA model takes the generated image and several questions as input and we take the probability of answering ``yes'' as the score for a question.
We compute the overall score by multiplying the probability of answering ``yes'' for each question.
The proposed disentangled BLIP-VQA is applied to evaluate the attribute binding for color, shape, and texture.

\subsection{UniDet-based Metric for Spatial Relationships and Numeracy Evaluation}
Many vision-language models exhibit limitations in reasoning spatial relationships, such as distinguishing between "left" and "right," as well as in numerical counting. Consequently, we propose the use of a detection-based evaluation metric to assess performance in spatial relationships and numeracy.

\textbf{2D-spatial relationships.}
We first use UniDet~\cite{zhou2022simple} to detect objects in the generated image.
Then we determine the spatial relationship between two objects by comparing the locations of the centers of the two bounding boxes. Denote the center of the two objects as $(x_1, y_1)$ and $(x_2, y_2)$, respectively. The first object is on the left of the second object if $x_1 < x_2, |x_1-x_2| > |y_1-y_2|$, and the intersection-over-union (IoU) between the two bounding boxes is below the threshold of $0.1$. Other spatial relationships ``right'', ``top'', and ``bottom'' are evaluated similarly.
We evaluate ``next to'', ``near'', and ``on the side of'' by comparing the distances between the centers of two objects with a threshold.

\textbf{3D-spatial relationships.}
For 3D-spatial relationships, we leverage depth estimation~\cite{ranftl2021vision}, in conjunction with the detection-based metric. Let $d_1$ and $d_2$ represent the mean values of the depth maps corresponding to the two objects. Specifically, if $d_1 > d_2$ and the IoU metric exceeds the predetermined threshold of 0.5, it is inferred that the first object is positioned in front of the second object. Other 3D-spatial relationships “behind”, “hidden by” are evaluated similarly.

\textbf{Numeracy.}
For numeracy, we first extract the names of objects and their corresponding quantities from the prompts. Then, we leverage UniDet to detect objects in the images. 
This scoring mechanism is designed to consider both objects and their numerical correctness proportionally based on the number of object categories mentioned in the prompt.
Let the variable $n$ represent the count of distinct object categories referenced in the given prompt. For every identified object in the image, the assigned score is calculated as $1/(2n)$ points. Furthermore, another $1/(2n)$ points are allocated if the generated quantity aligns accurately with the specified category.

\subsection{3-in-1 Metric for Complex Compositions Evaluation}
Since different evaluation metrics are designed for evaluating different types of compositionality, there is no single metric that works well for all categories. 
For non-MLLM metric, we empirically find that the Disentangled BLIP-VQA works best for attribute binding evaluation, UniDet-based metric works best for 2D/3D-spatial relationship and numeracy evaluation, and CLIPScore works best for non-spatial relationship evaluation.
Thus, we design a 3-in-1 evaluation metric which computes the average score of CLIPScore, Disentangled BLIP-VQA, and UniDet, as the evaluation metric for complex compositions.

\subsection{MLLM-based Evaluation Metric}
Multimodal Large Language Models enable users to instruct LLMs to analyze user-provided image inputs. By integrating the visual modality, MLLMs enhance the capability of language-only systems, providing them with new interfaces to address a variety of tasks. However, the multimodal abilities of MLLMs regarding the compositional text-to-image generation remain unclear. Within this context, we analyze the abilities of three types of MLLMs, namely MiniGPT-4, ShareGPT4V~\cite{chen2023sharegpt4v} and GPT-4V~\cite{yang2023dawn}, focusing on their performance in compositional problems.

\textbf{Evaluation methods of MLLMs.}
By aligning a pretrained visual encoder with a frozen large language model, multimodal large language models such as \textit{MiniGPT-4}~\cite{zhu2023minigpt4} have demonstrated great abilities in vision-language cross-modal understanding. 
But the current MiniGPT-4 model exhibit limitations such as inaccurate understanding of images and hallucination issues.
\textit{ShareGPT4V}~\cite{chen2023sharegpt4v} is a large-scale image-text dataset featuring 1.2 million detailed captions characterized by richness and diversity. %
\textit{GPT-4V}, a state-of-the-art MLLM, is developed on the foundation of the state-of-the-art LLM, GPT-4~\cite{achiam2023gpt}, and trained extensively on a large-scale dataset containing multimodal information~\cite{achiam2023gpt}. Employing MLLM as an evaluation metric, we submit generated images to the model and assess their alignment with the provided text prompt. This evaluation involves soliciting predictions for the image-text alignment score.

\textbf{Prompt template for MLLM evaluation.}
We leverage MLLMs as an evaluation metric by feeding the generated images to the model and asking two questions with Chain-of-Thought~\cite{wei2022chain}: ``describe the image'' and ``predict the image-text alignment score''. We detail the prompts used for MLLM evaluation metric. 
For each sub-category, we ask two questions in sequence: ``describe the image'' and ``predict the image-text alignment score''.
Specifically, Table~\ref{app:CoT_color_to_texture} shows the prompts for evaluating attribute binding (color, shape, texture). Similar to BLIP-VQA, for each noun phrase in a prompt, we request a number equivalent to the noun phrase and perform a multiplication. Table~\ref{app:CoT_spatial}, Table~\ref{app:CoT_numeracy}, Table~\ref{app:CoT_nonspatial}, and Table~\ref{app:CoT_complex} demonstrate the prompt templates used for 2D/3D-spatial relationships, non-spatial relationships, numeracy, and complex compositions, respectively\footnote{We empirically observed that, using a five-graded marking system instead of hundred-mark system enhances the performances of ShareGPT4V.}. Due to the robust capabilities of GPT-4V and limited quota constraints, we have omitted ``describe the image'' for prompt of GPT-4V.

For MLLM without Chain-of-Thought in addressing specific compositional problems, we utilize predefined prompts that prompt MLLM to provide a score ranging from 0 to 100. For attribute binding, we focus on the presence of specific objects and their corresponding attributes. We utilize a prompt template such as \textit{``Is there \{object\} in the image? Give a score from 0 to 100. If \{object\} is not present or if \{object\} is not \{color/shape/texture description\}, give a lower score.''} 
We leverage this question for each noun phrase in the text and compute the average score. 
For the spatial relationships, non-spatial relationships, and complex compositions, we employ a more general prompt template such as \textit{``Rate the overall alignment between the image and the text prompt \{prompt\}. Give a score from 0 to 100.''}.

\input{tables/tab_app_cot_attribute}
\input{tables/tab_app_cot_spatial}
\input{tables/tab_app_cot_nonspatial}
\input{tables/tab_app_cot_numeracy}
\input{tables/tab_app_cot_complex}

%% file: tables/tab_app_cot_attribute.tex
\begin{table}[h]
\caption{Prompts details for MLLM evaluation on attribute binding.}
\label{app:CoT_color_to_texture}
\resizebox{\linewidth}{!}{
\begin{tabular}{@{}ll@{}}
\toprule
Describe                   & \begin{tabular}[c]{@{}l@{}}You are my assistant to identify   any objects and their color (shape, texture) in the image. \\      Briefly describe what it is in the image within 50 words.\end{tabular}                                                                                                                                                                                                                                                                                                                                                        \\ \midrule
Predict & \begin{tabular}[c]{@{}l@{}}According to the image and your   previous answer, evaluate if there is \{adj.+noun\} in the image. \\      Give a score from 0 to 100, according the criteria:\\      100: there is \{noun\}, and \{noun\} is \{adj\}.\\      75: there is \{noun\}, \{noun\} is mostly \{adj\}.\\      20: there is \{noun\}, but it is not \{adj\}.\\      10: no \{noun\} in the image.\\      Provide your analysis and explanation in JSON format with the following   keys: score (e.g., 85), 
\\explanation (within 20 words).\end{tabular} \\ \bottomrule
\end{tabular}
}
\end{table}

%% file: tables/tab_app_cot_spatial.tex
\begin{table}[h]
\caption{Prompts details for MLLM evaluation on 2D/3D-spatial relationship.}
\label{app:CoT_spatial}
\resizebox{\linewidth}{!}{
\begin{tabular}{@{}ll@{}}
\toprule
Describe & \begin{tabular}[c]{@{}l@{}}You are my assistant to identify   objects and their spatial layout in the image. \\      Briefly describe the image within 50 words.\end{tabular}                                                                                                                                                                                                                                                                                                                                                                                                                                                                                                                             \\ \midrule
Predict  & \begin{tabular}[c]{@{}l@{}}According to the   image and your previous answer, evaluate if the text "\{xxx\}" is   correctly portrayed in the image.\\      Give a score from 0 to 100, according the criteria:\\      100: correct spatial layout in the image for all objects mentioned in the   text.\\      80: basically, spatial layout of objects matches the text.\\      60: spatial layout not aligned properly with the text.\\      40: image not aligned properly with the text.\\      20: image almost irrelevant to the text.\\      Provide your analysis and explanation in JSON format with the following   keys: score (e.g., 85), \\      explanation (within 20 words).\end{tabular} \\ \bottomrule
\end{tabular}
}
\end{table}

%% file: tables/tab_app_cot_nonspatial.tex
\begin{table}[h]
\caption{Prompts details for MLLM evaluation on non-spatial relationship.}
\label{app:CoT_nonspatial}
\resizebox{\linewidth}{!}{
\begin{tabular}{@{}ll@{}}
\toprule
Describe & \begin{tabular}[c]{@{}l@{}}You are my assistant to identify   the actions, events, objects and their relationships in the image.\\      Briefly describe the image within 50 words.\end{tabular}                                                                                                                                                                                                                                                                                                                                                                                                                                                                                                                                                                                                                                                                          \\ \midrule
Predict  & \begin{tabular}[c]{@{}l@{}}According to the image and your   previous answer, evaluate if the text "\{xxx\}" is correctly   portrayed in the image.\\      Give a score from 0 to 100, according the criteria:\\      100: the image accurately portrayed the actions, events and relationships   between objects described in the text.\\      80: the image portrayed most of the actions, events and relationships but   with minor discrepancies.\\      60: the image depicted some elements, but action relationships between   objects are not correct.\\      40: the image failed to convey the full scope of the text.\\      20: the image did not depict any actions or events that match the   text.\\      Provide your analysis and explanation in JSON format with the following   keys: score (e.g., 85),\\      explanation (within 20 words).\end{tabular} \\ \bottomrule
\end{tabular}
}
\end{table}

%% file: tables/tab_app_cot_numeracy.tex
\begin{table}[h]
\caption{Prompts details for MLLM evaluation on numeracy.}
\label{app:CoT_numeracy}
\resizebox{\linewidth}{!}{
\begin{tabular}{@{}ll@{}}
\toprule
Describe & \begin{tabular}[c]{@{}l@{}}You are my assistant to identify objects and their quantities in the image.\\      Briefly describe the image within 50 words.\end{tabular}                                                                                                                                                                                                                                                                                                                                                                                                                                                                                                                                                                                                                                                                          \\ \midrule
Predict  & \begin{tabular}[c]{@{}l@{}}According to the image and your   previous answer, evaluate how well the image aligns with the text prompt: "\{xxx\}".\\      Give a score from 0 to 100, according the criteria:\\      100:  correct numerical content in the image for all objects mentioned in the text.\\      80: basically, numerical content of objects matches the text.\\      60: numerical content not aligned properly with the text.\\      40: image not aligned properly with the text.\\      20: image almost irrelevant to the text.\\      Provide your analysis and explanation in JSON format with the following keys: score (e.g., 85),\\      explanation (within 20 words).\end{tabular} \\ \bottomrule
\end{tabular}
}
\end{table}

%% file: tables/tab_app_cot_complex.tex
\begin{table}[h]
\caption{Prompts details for MLLM evaluation on complex compositions.}
\label{app:CoT_complex}
\resizebox{\linewidth}{!}{
\begin{tabular}{@{}ll@{}}
\toprule
Describe & \begin{tabular}[c]{@{}l@{}}You are my assistant to evaluate   the correspondence of the image to a given text prompt.\\      Briefly describe the image within 50 words, focus on the objects in the   image and their attributes (such as color, shape, texture), 
\\spatial layout and   action relationships.\end{tabular}                                                                                                                                                                                                                                                                                                                                                                                                                                                                                                                              \\ \midrule
Predict  & \begin{tabular}[c]{@{}l@{}}According to the   image and your previous answer, evaluate how well the image aligns with the   text prompt: \{xxx\}.\\      Give a score from 0 to 100, according the criteria:\\      100: the image perfectly matches the content of the text prompt, with no discrepancies.\\      80: the image portrayed most of the actions, events and relationships but   with minor discrepancies.\\      60: the image depicted some elements in the text prompt, but ignored some   key parts or details.\\      40: the image did not depict any actions or events that match the   text.\\      20: the image failed to convey the full scope in the text prompt.\\      Provide your analysis and explanation in JSON format with the following   keys: score (e.g., 85), \\      explanation (within 20 words).\end{tabular} \\ \bottomrule
\end{tabular}
}
\end{table}

%% file: sections/method.tex
\section{Boosting Compositional Text-to-image Generation with GORS}
\label{method}
We introduce a simple but effective approach, \method~(\abbr), to improve the compositional ability of pretrained text-to-image models. Our approach finetunes a pretrained text-to-image model such as Stable Diffusion~\cite{rombach2022high} with generated images that highly align with the compositional prompts, where the fine-tuning loss is weighted by the reward which is defined as the alignment score between compositional prompts and generated images.

Specifically, given the text-to-image model $p_\theta$ and a set of text prompts $y_1, y_2, \cdots, y_n$, we first generate $k$ images for each text prompt, resulting in $kn$ generated images $x_1, x_2, \cdots, x_{kn}$.
Text-image alignment scores $s_1, s_2, \cdots, s_{kn}$ are predicted as rewards.
We select the generated images whose rewards are higher than a threshold to fine-tune the text-to-image model. The selected set of samples are denoted as $\mathcal{D}_s$.
During fine-tuning, we weight the loss with the reward of each sample. Generated images that align with the compositional prompt better are assigned higher loss weights, and vice versa.
The loss function for fine-tuning is
\begin{equation}
    \mathcal{L}(\theta)=\mathbb{E}_{(x,y,s)\in \mathcal{D}_s}{\large\left[s \cdot\left\|\epsilon-\epsilon_\theta\left(z_t, t, y\right)\right\|_2^2\large\right]},
\end{equation}
where $(x, y, s)$ is the triplet of the image, text prompt, and reward, and $z_t$ represents the latent features of $x$ at timestep $t$.
We adopt LoRA~\cite{hu2021lora} for efficient finetuning.

%% file: sections/experiments.tex
\section{Experiments}
\label{others}
\subsection{Experimental Setup}

\textbf{Evaluated models.}
We evaluate the performance of 11 text-to-image models on \name: 
\textit{Stable Diffusion v1-4}, \textit{Stable Diffusion v2}~\cite{rombach2022high}, \textit{Composable Diffusion}~\cite{liu2022compositional}, \textit{Structured Diffusion}~\cite{feng2022training}, \textit{Attend-and-Excite}~\cite{chefer2023attend}, and\textit{~\abbr}. 
We further conduct evaluations on 5 state-of-the-art text-to-image models: \textit{PixArt-$\alpha$}~\cite{chen2024pixart}, \textit{Stable Diffusion XL}~\cite{podell2023sdxl}, \textit{DALL-E 3}~\cite{betker2023improving}, \textit{Stable Diffusion 3}~\cite{esser2024scaling} and \textit{FLUX.1}~\cite{FLUX}. 
\textit{\abbr} is our proposed approach which finetunes Stable Diffusion v2 with selected samples and their rewards. 

\textit{Stable Diffusion v1-4} and \textit{Stable Diffusion v2}~\cite{rombach2022high} are text-to-image models trained on large amount of image-text pairs.
\textit{Composable Diffusion}~\cite{liu2022compositional} is designed for conjunction and negation of concepts for pretrained diffusion models.
\textit{Structured Diffusion}~\cite{feng2022training} and \textit{Attend-and-Excite}~\cite{chefer2023attend} are designed for attribute binding for pretrained diffusion models. We re-implement those approaches on Stable Diffusion v2 to enable fair comparisons.
PixArt-$\alpha$~\cite{chen2024pixart} is a Transformer-based T2I diffusion model, boasting competitive image generation quality comparable to state-of-the-art image generators with low training costs. We test PixArt-$\alpha$ fine-tuned on~\abbr, denoted as PixArt-$\alpha$-ft. Stable Diffusion XL 1.0~\cite{podell2023sdxl} building upon previous iterations of Stable Diffusion models, represents a powerful text-to-image generation framework. DALLE-3~\cite{betker2023improving} is a new text-to-image generation system, enhancing the alignment between generated images and provided text description. Stable Diffusion 3~\cite{esser2024scaling} combines a diffusion transformer architecture and flow matching. FLUX.1 [schnell]~\cite{FLUX} is a 12 billion parameter rectified flow transformer.

To avoid the bias from selecting samples by evaluation metrics as reward~\cite{park2021benchmark, teney2020value}, we introduce new reward models which are different from our proposed evaluation metrics, denoted as \textit{~\abbr-unbiased}. Specifically, we adopt Grounded-SAM~\cite{liu2023grounding} as the reward model for the attribute binding category. We extract the segmentation masks of attributes and their associated nouns separately with Grounded-SAM, and use the Intersection-over-Union (IoU) between the attribute masks and the noun masks together with the grounding mask confidence to represent the attribute binding performance. We apply GLIP-based~\cite{li2022grounded} selection method for 2D/3D-spatial relationships and numeracy. For non-spatial relationships, we adopt BLIP~\cite{li2022blip} to generate image captions and CLIP~\cite{radford2021learning, hessel2021clipscore} to measure the text-text similarity between the generated captions and the input text prompts. For complex compositions, we integrate the 3 aforementioned reward models as the total reward. Those sample selection models are different from the models used as evaluation metrics.

\textbf{Implementation details.} We employ pre-trained models for our proposed evaluation metrics. For BLIP-VQA, we utilize the BLIP w/ ViT-B and CapFilt-L~\cite{li2022blip} pretrained on image-text pairs and fine-tuned on VQA. We employ the UniDet~\cite{zhou2022simple} model
trained on 4 large-scale detection datasets (COCO~\cite{lin2014microsoft}, Objects365~\cite{shao2019objects365}, OpenImages~\cite{kuznetsova2020open}, and Mapillary~\cite{neuhold2017mapillary}). For CLIPScore, we use the ``ViT\--{}B\//32'' pretrained CLIP model~\cite{hessel2021clipscore, radford2021learning}. For MiniGPT4-CoT, we utilize the Vicuna 13B of MiniGPT4~\cite{zhu2023minigpt4} variant with a temperature setting of 0.7 and a beam size of 1. For ShareGPT4V~\cite{chen2023sharegpt4v} and GPT-4V~\cite{yang2023dawn}, we use the default parameters, with temperature setting of 0.2 and 1, respectively.

We implement our proposed~\abbr~upon the codebase of diffusers~\cite{TP-toolbox-web} (Apache License), and finetune the self-attention layers of the CLIP text encoder and the attention layers of U-net using LoRA~\cite{hu2021lora}.
The model is trained by AdamW optimizer~\cite{loshchilov2019decoupled} with $\beta_1$=0.9, $\beta_2$=0.999, $\epsilon$=1\textit{e}-8, and weight decay of 0.01. The batch size is 5. The model is trained on 8 32GB NVIDIA v100 GPUs.

\subsection{Evaluation Metrics}

We generate 10 images for each text prompt in T2I-CompBench for automatic evaluation. To ensure a fair comparison, the images are generated using the fixed seed across all models\footnote{Because the DALLE 3 API does not allow specifying random seed and due to the high cost of the DALLE 3 API, we use DALLE 3 to generate 3 images for each prompt without specified seed.}.

\textbf{Previous metrics.} \textit{CLIPScore}~\cite{radford2021learning,hessel2021clipscore} (denoted as \textit{CLIP}) calculates the cosine similarity between text features and generated-image features extracted by CLIP.
\textit{BLIP-CLIP}~\cite{chefer2023attend} (denoted as \textit{B-CLIP})
applies BLIP~\cite{li2022blip} to generate captions for the generated images, and then calculates the CLIP text-text cosine similarity between the generated captions and text prompts. \textit{BLIP-VQA-naive} (denoted as \textit{B-VQA-n}) applies BLIP VQA to ask a single question (e.g., a green bench and a red car?) with the whole prompt.

\textbf{Our proposed metrics.}
\textit{Disentangled BLIP-VQA} (denoted as \textit{B-VQA}) is our proposed evaluation metric for attribute binding.
\textit{UniDet} is our proposed metric for 2D/3D-spatial relationships and numeracy evaluation metric.

For evaluating MLLM to serve as a metric, we assess three models, and propose MLLM-based metric for non-spatial relationship and complex compositions: \textit{MiniGPT4} (denoted as \textit{mGPT}), \textit{ShareGPT4V} (denoted as \textit{Share}), and \textit{GPT-4V}. Due to the restricted quota for GPT-4V, we conducted evaluations on one-fifth of the images of~\name~(\textit{i.e.}, 600 images for each category). To boost capability of MLLM, we apply Chain-of-Thought~\cite{wei2022chain} (denoted as \textit{-CoT}) for mGPT and ShareGPT4V.

\begin{figure*}
  \begin{minipage}{0.32\linewidth}
    \includegraphics[width=\linewidth]{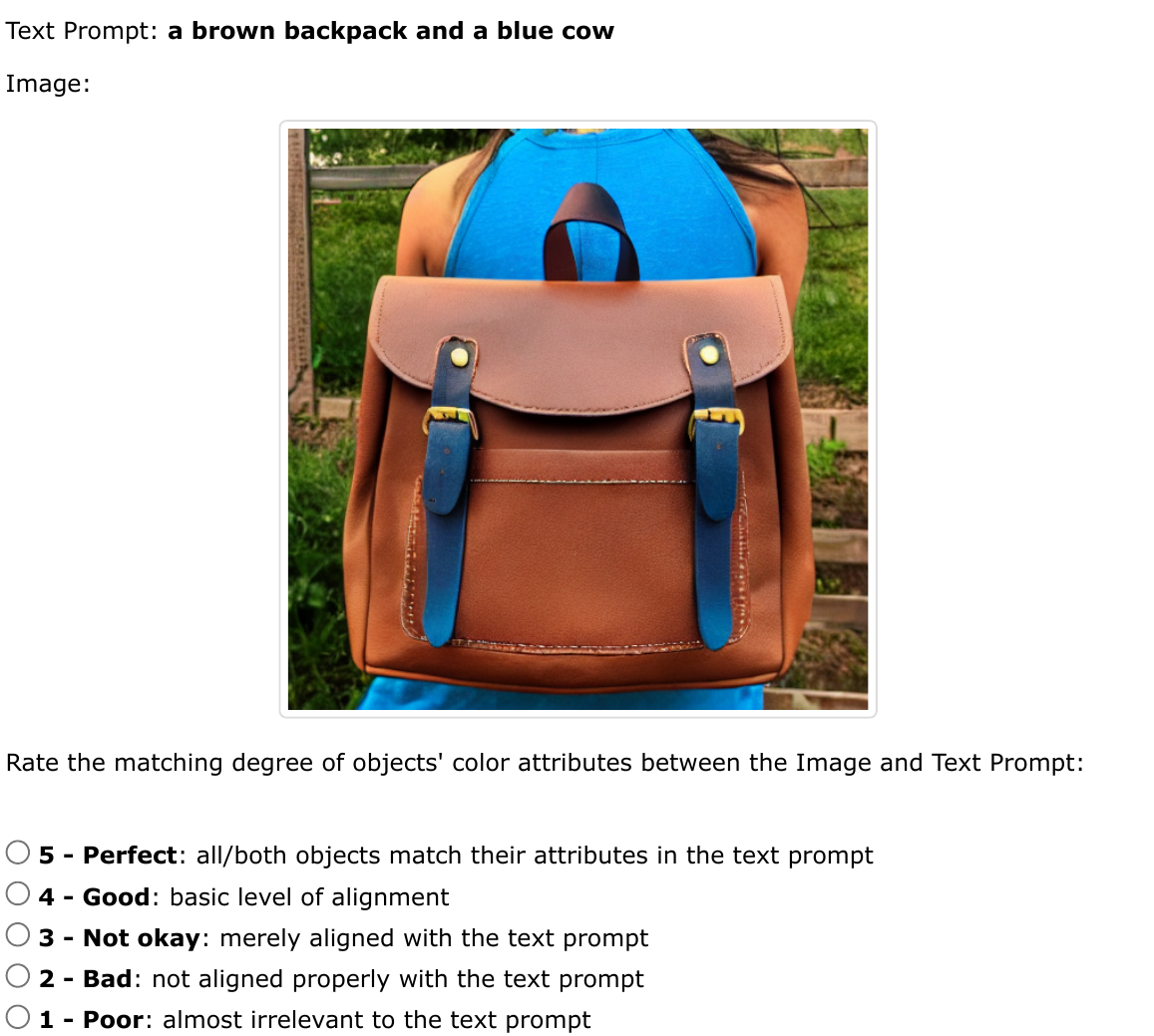}
  \end{minipage}
  \begin{minipage}{0.32\linewidth}
    \includegraphics[width=\linewidth]{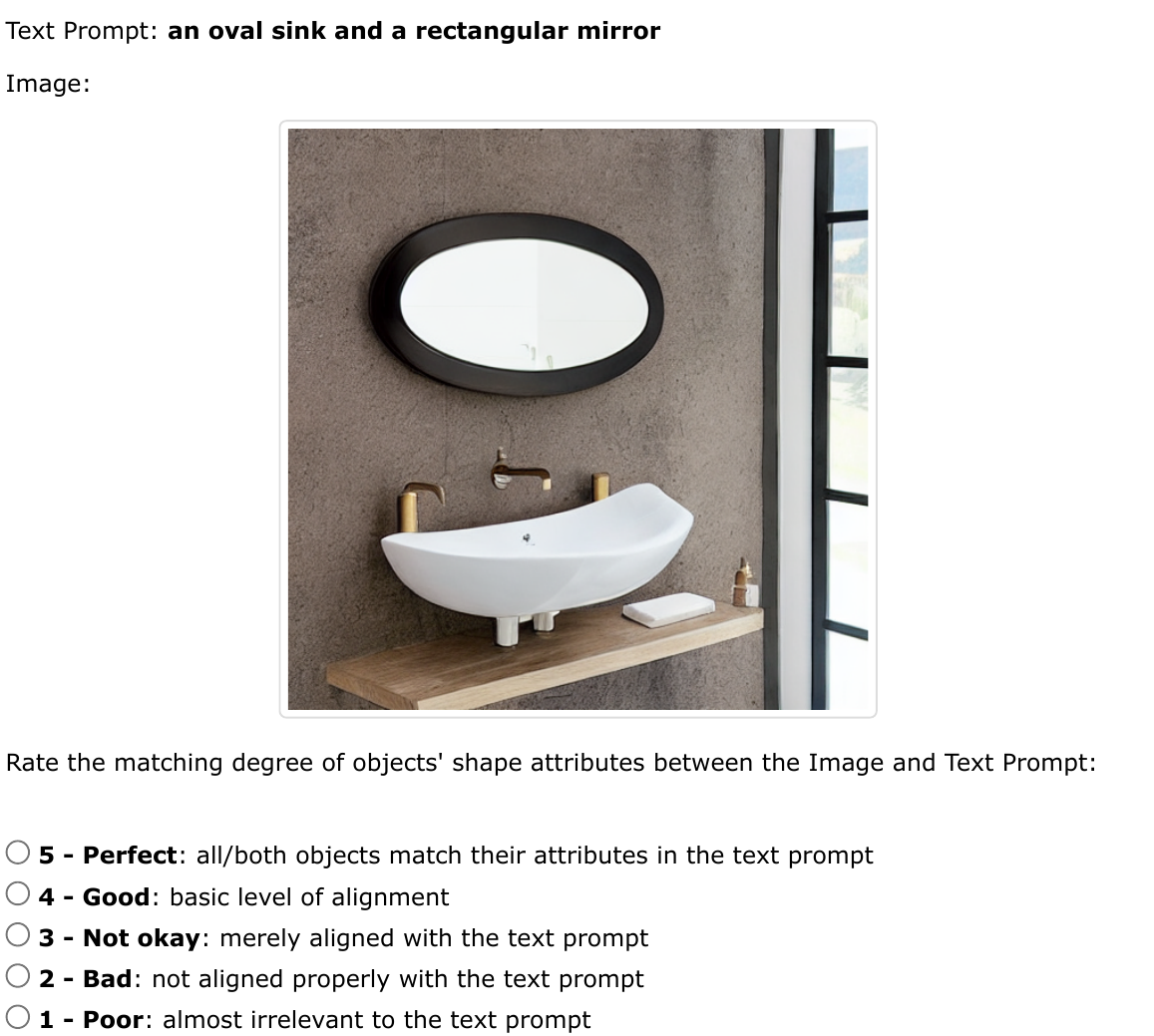}
  \end{minipage}
  \begin{minipage}{0.32\linewidth}
    \includegraphics[width=\linewidth]{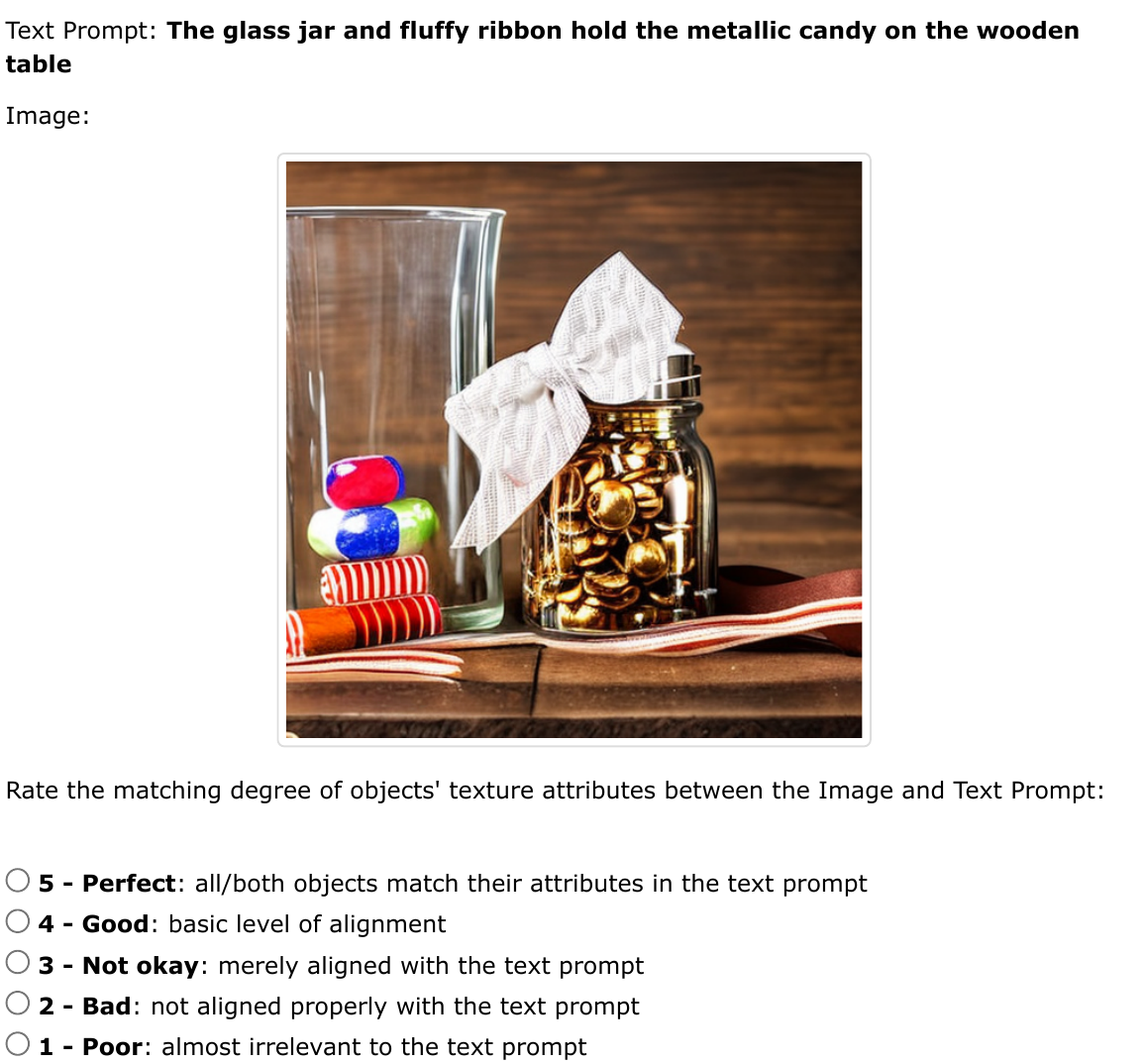}
  \end{minipage}
  \caption{AMT interface for the image-text alignment evaluation on attribute binding (color, shape, texture).}\label{fig:AMT1}
\end{figure*}

\begin{figure*}
  \begin{minipage}{0.32\linewidth}
    \includegraphics[width=\linewidth]{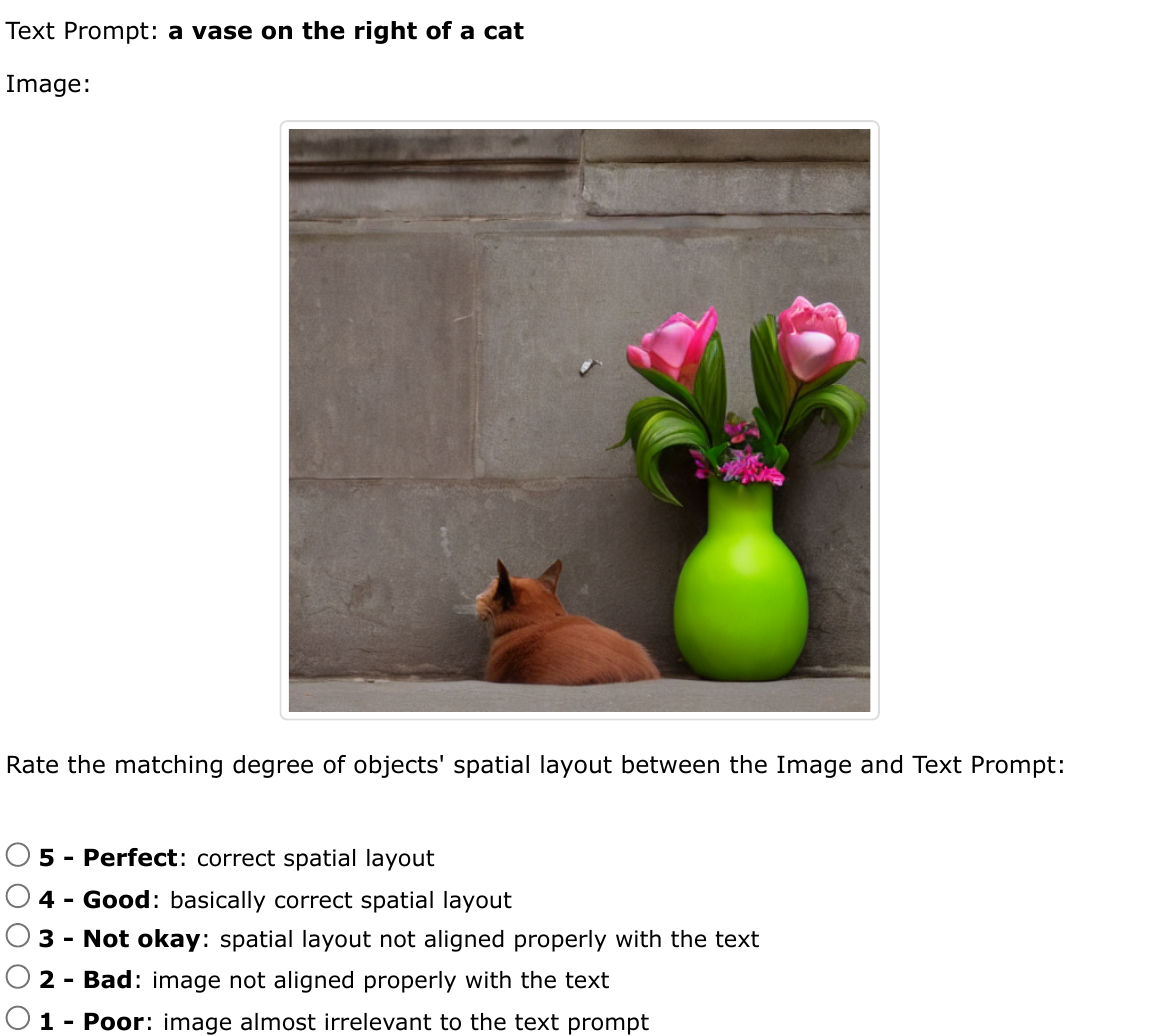}
  \end{minipage}
  \begin{minipage}{0.32\linewidth}
    \includegraphics[width=\linewidth]{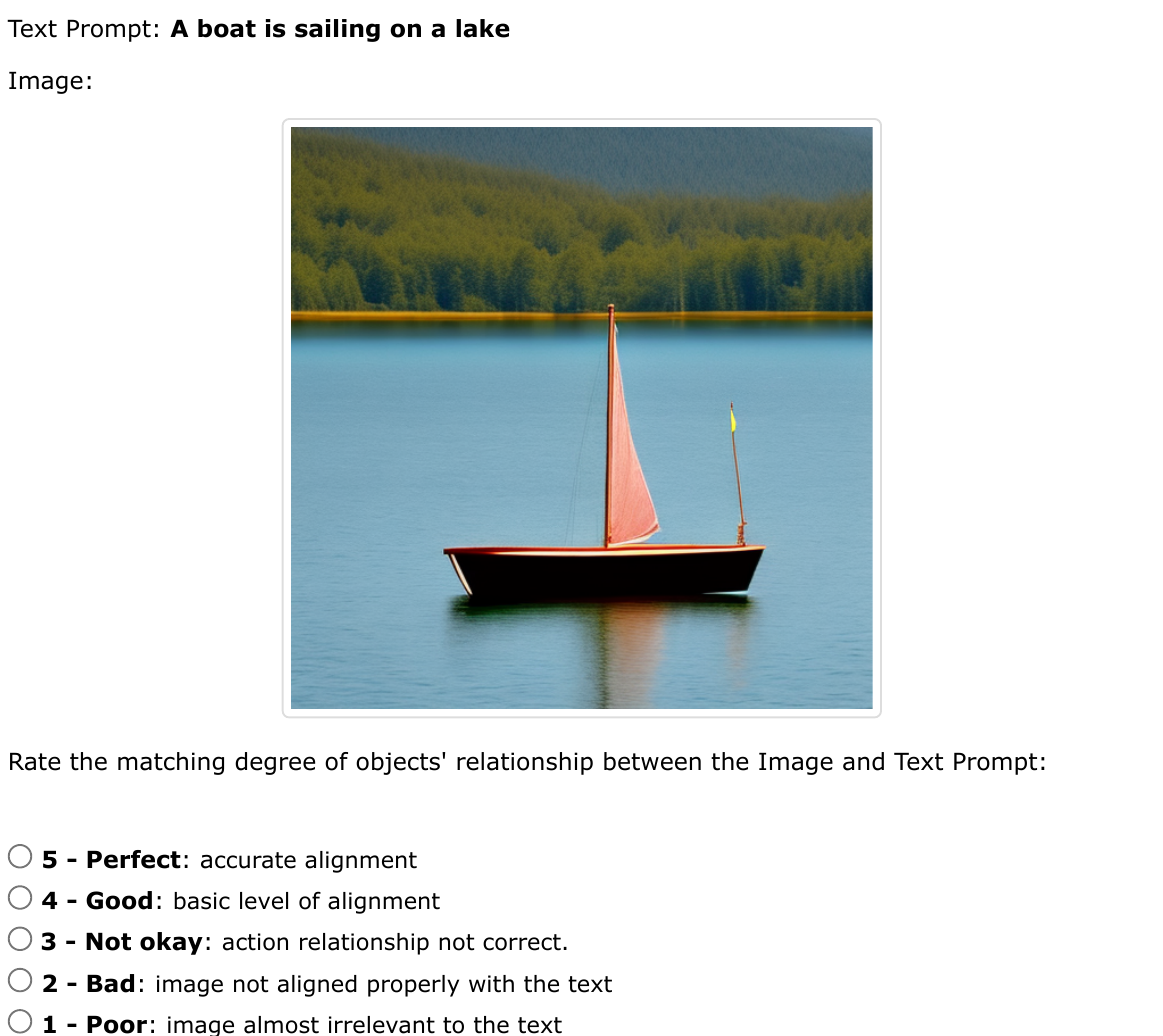}
  \end{minipage}
  \begin{minipage}{0.32\linewidth}
    \includegraphics[width=\linewidth]{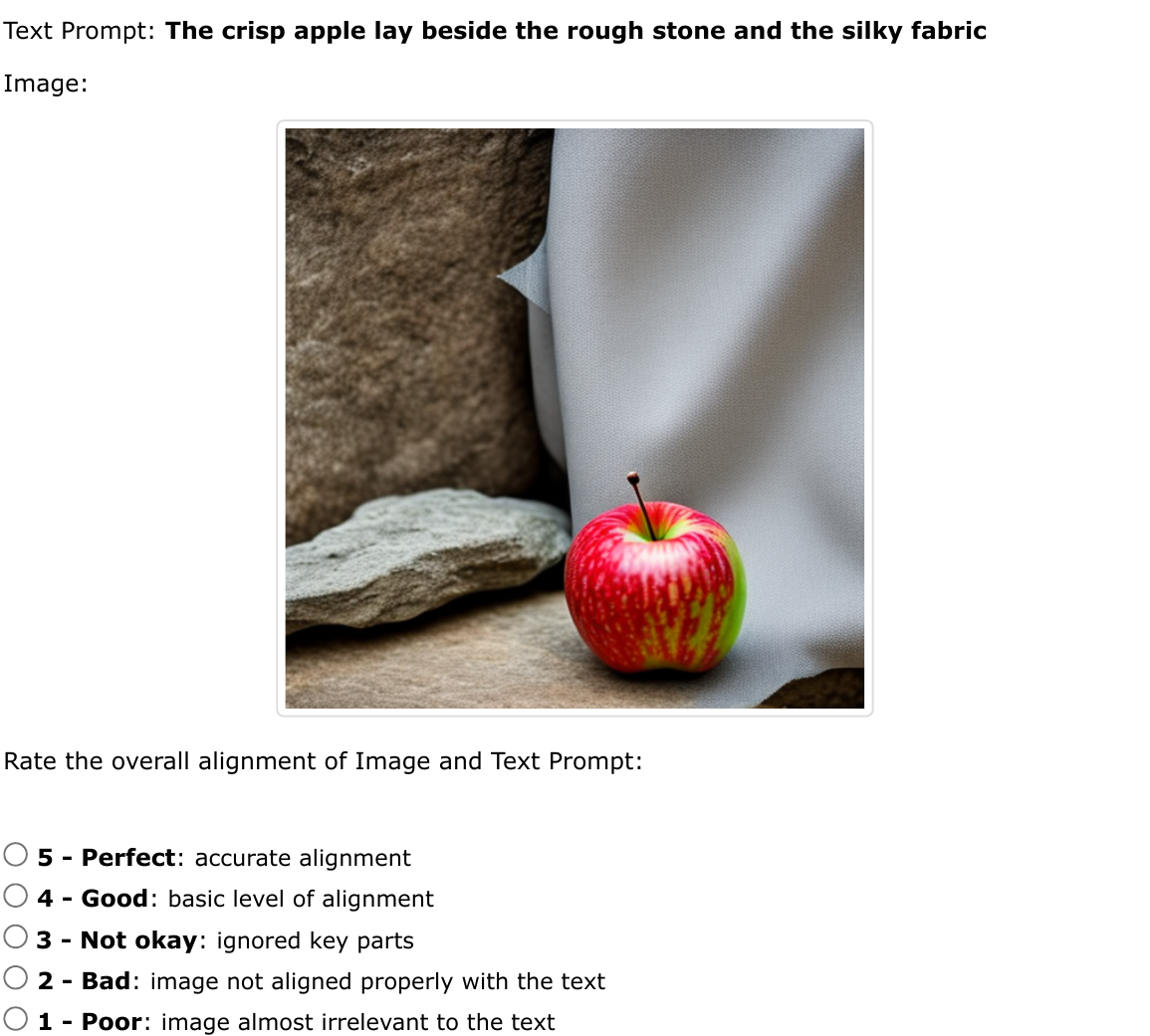}
  \end{minipage}
  \caption{AMT interface for the image-text alignment evaluation on 2D-spatial relationships, non-spatial relationships, and complex compositions.}\label{fig:AMT2}
\end{figure*}

\begin{figure*}
\centering
\includegraphics[width=0.63\linewidth]{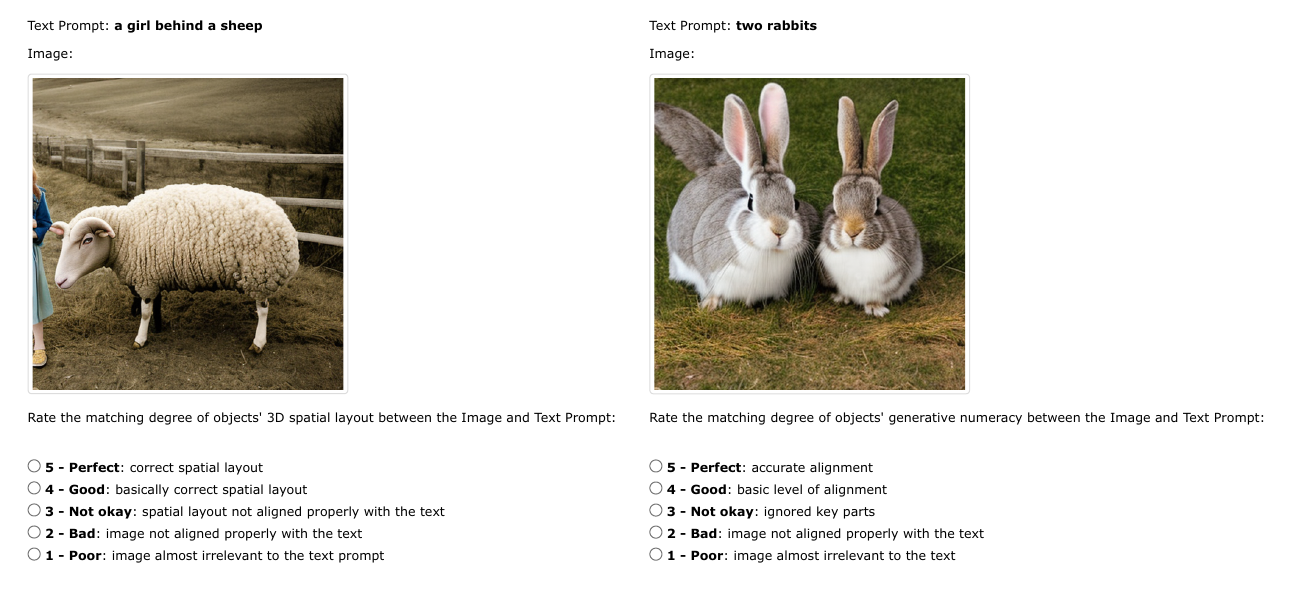}
  \caption{AMT interface for the image-text alignment evaluation on 3D-spatial relationships, and generative numeracy.}\label{fig:AMT3}
\end{figure*}

\textbf{Human evaluation.}
For human evaluation of each sub-category, we randomly select 25 prompts and generate 2 images per prompt, resulting in 300 images generated with 200 prompts per model in total. The testing set includes 300 prompts for each sub-category, resulting in 2,400 prompts in total. The prompt sampling rate for human evaluation is $8.33\%$. We utilize Amazon Mechanical Turk and ask three workers to score each generated-image-text pair independently based on the image-text alignment. %
The worker can choose a score from $\{1,2,3,4,5\}$, as shown in Figure~\ref{fig:AMT1}-\ref{fig:AMT3}. We normalize the scores by dividing them by 5. We then compute the average score across all images and all workers.

\input{tables/tab_color_shape_test}

\input{tables/tab_texture_spatial}
\input{tables/tab_num_3d}
\input{tables/tab_non_spatial_compre}
\input{tables/tab_correlation}

\input{tables/tab_new_models}

\subsection{Comparison across Different Evaluation Metrics}

\textbf{Comparions of different evaluation metrics and human evaluation.} We show the results of different evaluation metrics in Table~\ref{benchmark:color_shape}-\ref{benchmark:non_spatial_compre}. Previous evaluation metrics, CLIP and B-CLIP, predict similar scores across different models and cannot reflect the differences between models. Our proposed metrics (highlighted in blue column) show a similar trend with human evaluation (highlighted in green column).

\textbf{Human correlation of the evaluation metrics.} We calculate Kendall's tau ($\tau$) and Spearman's rho ($\rho$) to evaluate the ranking correlation between automatic evaluation and human evaluation. 
The human correlation results are illustrated in Table~\ref{human_corr}.
The results verify the effectiveness of our proposed evaluation metrics, BLIP-VQA for attribute binding, UniDet-based metric for spatial relationships and numeracy.
For MLLM evaluation, MiniGPT4 does not perform well in terms of correlation with human perception. Share-CoT and GPT-4V excel in terms of non-spatial and complex categories, followed by CLIPScore and 3-in-1 evaluation metrics. In shape attribute, they exhibit performances that are comparable to non-MLLM metrics. In color, texture, 2D/3D-spatial and numeracy categories, their performances are slightly lower than non-MLLM metrics.

\textbf{Conclusion on the evaluation metrics.} Based on the human correlation of different evaluation metrics, we draw the conclusion for the optimal evaluation metrics for each category or sub-category, and those optimal metrics are used to benchmark different text-to-image generation methods in the following subsections. (1) For attribute binding, the best evaluation metric is BLIP-VQA. (2) For 2D spatial relationships, 3D spatial relationships, and numeracy, the UniDet-based metric performs best.
(3) For non-spatial relationships, the best metric is GPT-4V, the second-best metric is Share-CoT (ShareGPT4V with Chain-of-Thought), and the best non-MLLM metric is CLIP.
(4) For complex compositions, the best metric is GPT-4V, the second-best metric is Share-CoT (ShareGPT4V with Chain-of-Thought), and the best non-MLLM metric is 3-in-1. 

\subsection{Discussion about MLLMs as a unified metric}\label{MLLM part}
\textbf{Comparisons among different MLLMs.} We compare the human correlation among MiniGPT4, ShareGPT4V and GPT-4V. MiniGPT4 shows inadequate correlation with human perception\footnote{To facilitate comparisons of different MLLMs' capabilities, we consider the same evaluation metric as a unified entity. For instance, mGPT and mGPT-CoT are treated as a single entity (the same for Share and Share-CoT), and we compare them based on the highest human correlation values attained.}.
ShareGPT4V and GPT-4V excel in terms of non-spatial and complex categories. In color, texture, 2D/3D-spatial and numeracy categories, their performances are slightly lower than non-MLLM metrics. 
ShareGPT4V ranks second in non-spatial relationship and complex, ranging between first to third place in attribute categories, third in 2D-spatial, third (in terms of $\tau$) and second (in terms of $\rho$) in 3D-spatial, second in numeracy.
GPT-4V ranks first in non-spatial relationship and complex, second in color attribute, third in shape and texture attributes, second in 2D-spatial, second (in terms of $\tau$) and third (in terms of $\rho$) in 3D-spatial, third in numeracy.

\textbf{Comparisons of the effectiveness of Chain-of-Thought for MLLMs.} 
Utilizing Chain-of-thought~\cite{wei2022chain} to stimulate MLLM’s evaluation ability, we conduct the comparisons on MiniGPT4 and ShareGPT4V. With CoT, both MiniGPT4 and ShareGPT4V demonstrate improvements across most categories in terms of human correlation, highlighting the significance of effective prompting. 

We benchmark mGPT without Chain-of-Thought in Table~\ref{app:mGPT}, which shows the additional results of benchmarking on \name~of 6 models with MiniGPT-4 without Chain-of-Thought. Results indicate that MiniGPT-4 evaluation without Chain-of-Thought does not strictly align with human evaluation results.

\input{tables/tab_minigpt4}

\textbf{Discussion about the stability of MLLMs' results.}
We test the stability of the MLLM metric by conducting multiple executions on a consistent set of images. Specifically, we employ ShareGPT4V and GPT-4V to analyze 50 images from Stable v2 color category, repeating the process five times. 
Default parameters were used, with a temperature of 1.0 for GPT-4V and 0.2 for ShareGPT4V. 
As shown in Table~\ref{stability}, our analysis consistently produces results, with variations remaining within 0.032 across the five executions for GPT-4V and 0.0273 for ShareGPT4V. The average standard deviations are 7.3235 for GPT-4V and 6.8741 for ShareGPT4V.

\textbf{Limitations of MLLMs as a metric.} 
Share-CoT and GPT-4V have limits despite their strong performance as evaluation metrics. They do not adhere to the prompts of grading guidelines very well. Our empirical observations show that Share-CoT tends to give less diverse ratings, regardless of the prompt provided. GPT-4V can comprehend the evaluation prompts, but it is less able to convert into exact grades.

\subsection{Benchmarking on Different Methods}
\begin{figure}[!htb]
\centering
\includegraphics[width=\linewidth]{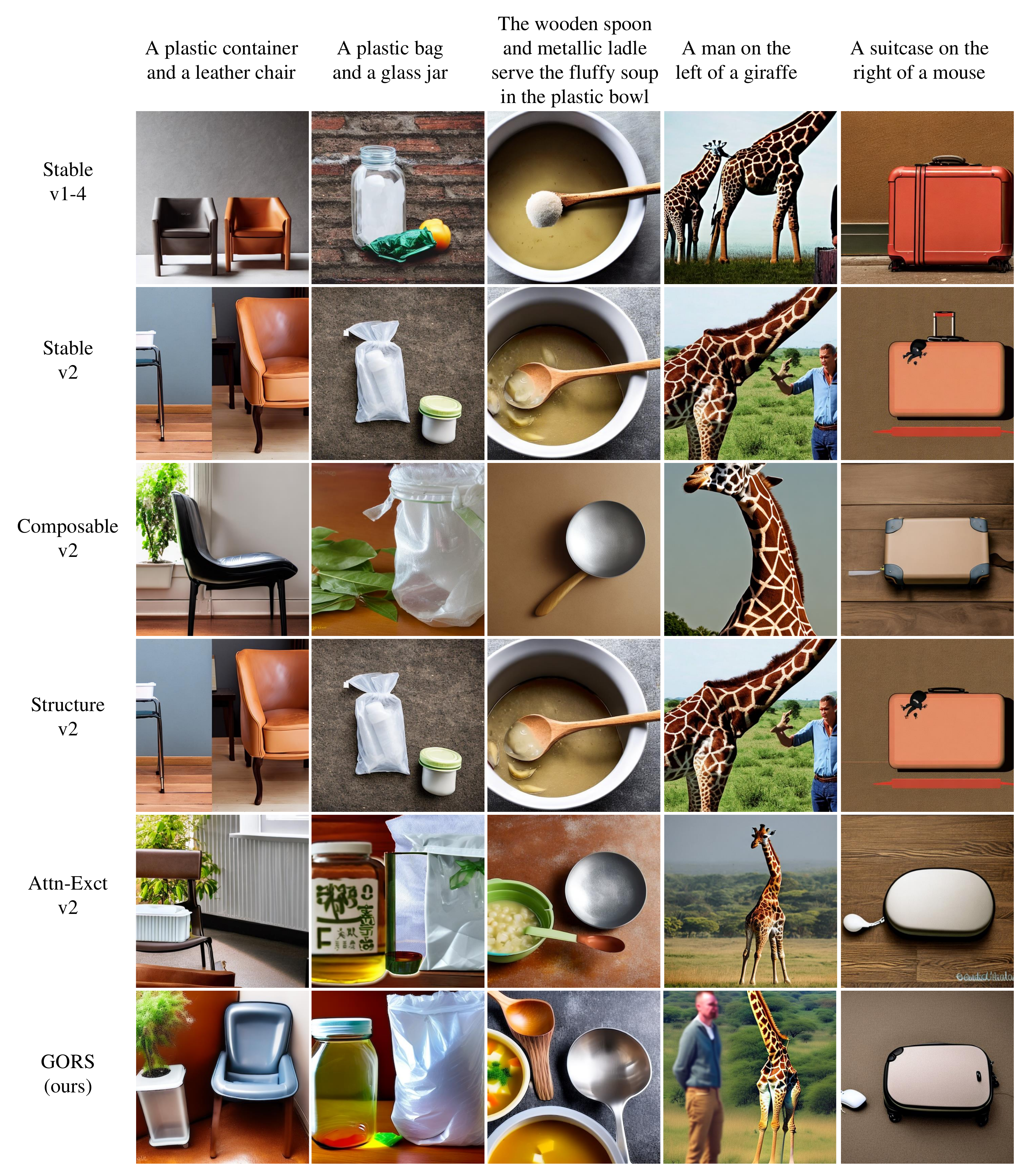}
\caption{Qualitative comparison between our approach and previous methods.}
\label{fig:results2}
\end{figure}

The quantitative benchmarking results are reported in Table~\ref{benchmark:color_shape}-\ref{benchmark:non_spatial_compre} and~\ref{benchmark:new_model}.
Qualitative results are shown in appendix Figure~\ref{fig:qualitat}, Figure~\ref{fig:results1} and Figure~\ref{fig:results2}.

\textbf{Comparisons across text-to-image models.}
(1) Stable Diffusion v2 consistently outperforms Stable Diffusion v1-4 in all types of compositional prompts and evaluation metrics.
(2) Although Structured Diffusion built upon Stable Diffusion v1-4 shows great performance improvement in attribute binding as reported in Feng~\textit{et al.}~\cite{feng2022training}, Structured Diffusion built upon Stable Diffusion v2 only brings slight performance gain upon Stable Diffusion v2. It indicates that boosting the performance upon a better baseline of Stable Diffusion v2 is more challenging.
(3) Composable Diffusion built upon Stable Diffusion v2 does not work well. A similar phenomenon was also observed in previous work~\cite{chefer2023attend} that Composable Diffusion often generates images containing a mixture of the subjects. %
In addition, Composable Diffusion was designed for concept conjunctions and negations so it is reasonable that it does not perform well in other compositional scenarios.
(4) Attend-and-Excite built upon Stable Diffusion v2 improves the performance in attribute binding.
(5) Previous methods Composable Diffusion~\cite{liu2022compositional}, Structure Diffusion~\cite{feng2022training} and Attend-and-Excite~\cite{chefer2023attend} are designed for concept conjunction or attribute binding, so they do not result in significant improvements in object relationships.
(6) Our proposed approach, \abbr, outperforms previous approaches across all types of compositional prompts, as demonstrated by the automatic evaluation, human evaluation, and qualitative results. The evaluation results of GORS-unbiased and GORS significantly exceed the baseline Stable v2. 
Besides, the performances of GORS-unbiased indicate
that our proposed approach is insensitive to the reward model used for selecting samples, and that the proposed approach works well as long as high-quality samples are selected. %
(7) Recent text-to-image models, Stable Diffusion XL~\cite{podell2023sdxl}, Pixart-$\alpha$-ft~\cite{chen2024pixart}, DALLE 3~\cite{betker2023improving}, Stable Diffusion 3~\cite{esser2024scaling} and FLUX.1 [schnell]~\cite{FLUX} improves across all categories compared to previous methods. Notably, DALLE 3 and SD3 achieves the state-of-the-art (SOTA) performance by a significant margin in almost all categories.

\textbf{Comparisons across compositionality categories.}
According to the human evaluation results, spatial relationship is the most challenging sub-category for text-to-image models, and attribute binding (shape) is also challenging.
Non-spatial relationship is the easiest sub-category.

\textbf{Comparisons between seen and unseen splits.}
We provide the seen and unseen splits for the test set in attribute binding, where the unseen set consists of attribute-object pairs that do not appear in the training set.
The unseen split tends to include more uncommon attribute-object combinations than seen split.
The performance comparison of seen and unseen splits for attribute binding is shown in Table~\ref{app:seen_unseen}.
Our observations reveal that our model exhibits slightly lower performance on the unseen set than the seen set.
\input{tables/tab_app_seen_unseen}

\textbf{Comparisons of the scalability of our proposed approach.}
To demonstrate the scalability of our proposed approach, we introduce additional 700 prompts of complex compositions to form an extended training set of 1,400 complex prompts. The new prompts are generated with the same methodology
. We conduct 6 experiments to train the models with different training set sizes, i.e., 25 prompts, 275 prompts, 350 prompts, 700 prompts, 1050 prompts, and 1400 prompts.
The results in Table~\ref{tab:add training data} show the performance of our model grows with the increase of the training set sizes. The results indicate the potential to achieve better performance by scaling up the training set. %

\input{tables/tab_add_training_data}

\begin{figure}[!htb]
    \includegraphics[width=\linewidth]{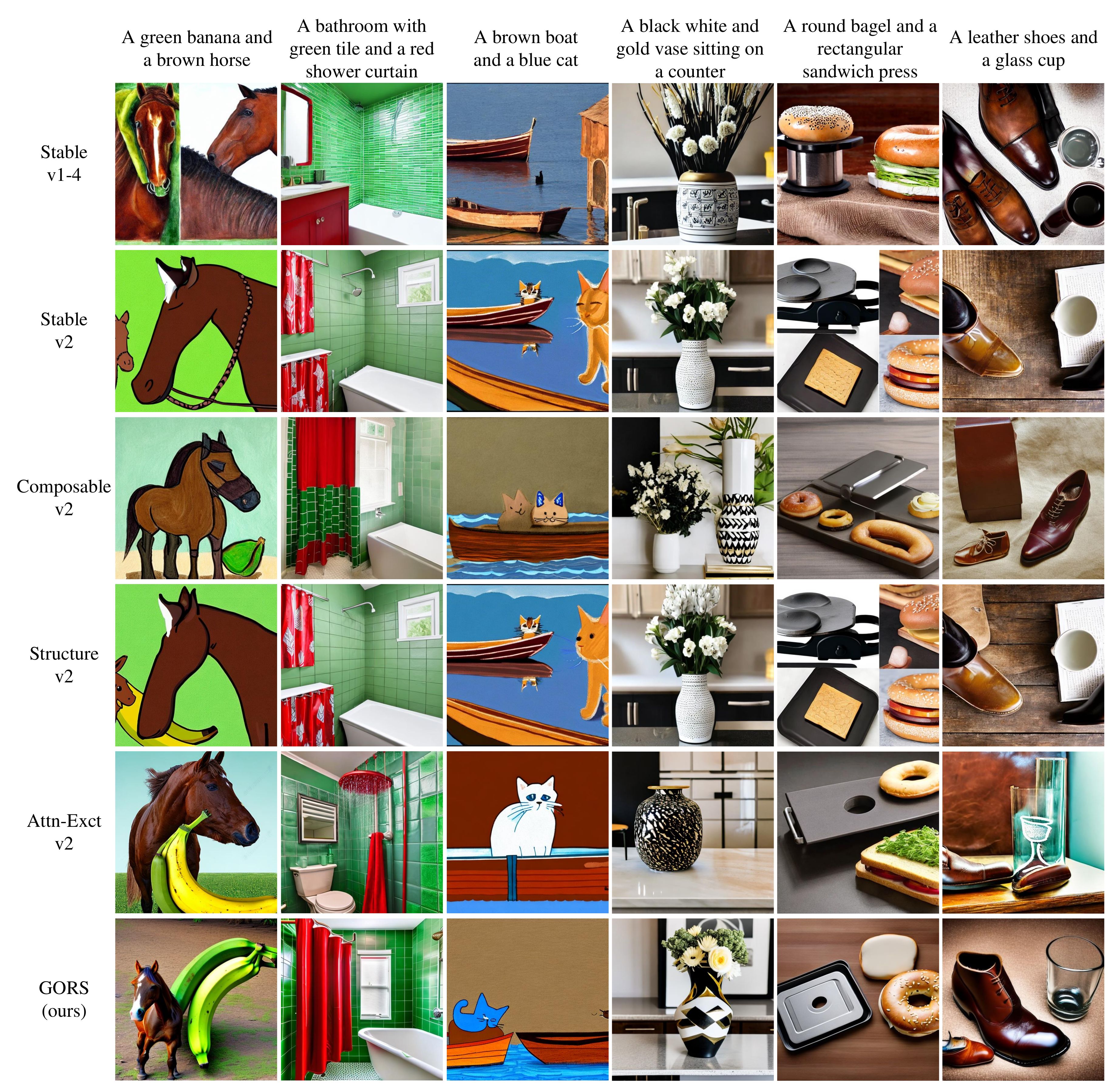}
    \caption{Qualitative comparison between our approach and previous methods.}
    \label{fig:qualitat}
\end{figure}

\begin{figure}[!htb]
\centering
\includegraphics[width=\linewidth]{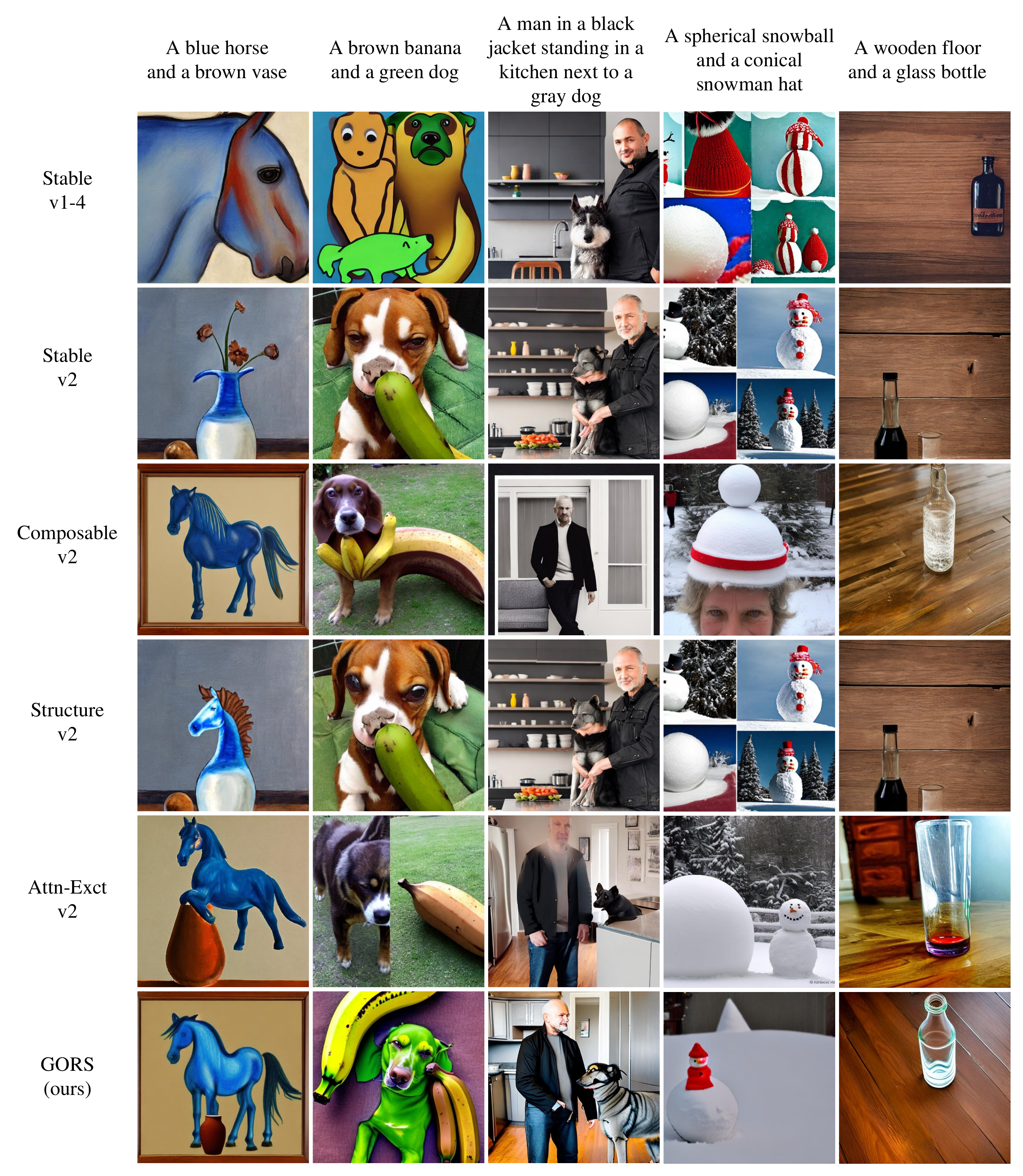}
\caption{Qualitative comparison between our approach and previous methods.}
\label{fig:results1}
\end{figure}

\textbf{Comparisons between original prompts and rephrased detailed prompts.}
Enhancing prompt details does not significantly improve the generative outcomes, as the challenges primarily lie in the visual content composition instead of the granularity of prompts. While more detailed prompts can improve image quality by providing additional descriptions, they do not necessarily enhance alignment between the generated images and the prompts.
To illustrate this, we use GPT-4 to rephrase the prompts to more detailed versions and compare the evaluation results. Specifically, we examine SDXL with attribute binding categories (e.g., color) as an example. The results, shown in Table~\ref{tab:blip_vqa_scores}, indicate that T2I models face similar compositional challenges with both original and detailed prompts.

\begin{table}[htb]
    \centering
    \caption{Comparison of BLIP-VQA scores between original and rephrased detailed prompts.}
    \label{tab:blip_vqa_scores}
    \resizebox{\linewidth}{!}{
    
    \begin{tabular}{lcc}
    
    \toprule
        & \makecell{\textbf{Original prompts} \\
        (avg length: 8.21 words)} & \makecell{\textbf{Rephrased detailed prompts} \\ (avg length: 24.00 words)} \\ \midrule
    B-VQA & 0.5879 & 0.5781 \\
    \bottomrule
    \end{tabular}
   } 
\end{table}

\subsection{Ablation Study}
We conduct ablation studies on our proposed \abbr~approach and evaluation metric.

\input{tables/tab_stability}

\begin{table*}[t]
\centering
\caption{Ablation studies on fine-tuning strategy and threshold.}
\label{ablation}
\small
\begin{tabular}{cccccc}
    \toprule
    \textbf{Metric}    & \textbf{FT U-Net only} & \textbf{FT CLIP only} & \textbf{Half threshold}  & \textbf{0 threshold} & \textbf{\abbr~(ours)}     \\
    \midrule
    B-VQA & 0.6216   & 0.5507   & 0.6157 & 0.6130& \textbf{0.6570}     \\
    \bottomrule
\end{tabular}
\end{table*}

\textbf{Finetuning strategy.}
Our approach finetunes both the CLIP text encoder and the U-Net of Stable Diffusion with LoRA~\cite{hu2021lora}. We conduct with the attribute binding (color) sub-category. We investigate the effects of finetuning CLIP only and U-Net only with LoRA.
As shown in Table~\ref{ablation}, our model which finetunes both CLIP and U-Net performs better.

\textbf{Threshold of selecting samples for finetuning.}
Our approach fine-tunes Stable Diffusion v2 with the selected samples that align well with the compositional prompts.
We manually set a threshold for the alignment score to select samples with higher alignment scores than the threshold for fine-tuning.
We experiment with setting the threshold to half of its original value, and setting the threshold to 0 (\textit{i.e.}, use all generated images for finetuning with rewards, without selection). Results in Table~\ref{ablation} demonstrate that half threshold and zero threshold will lead to worse performance.

\textbf{Threshold of UniDet-based metrics.} The threshold for the UniDet-based metric is validated via user studies with human correlation, as shown in Table~\ref{tab:unidet_correlation}. Among various tested thresholds (0.0, 0.25, 0.50, 0.75), the chosen threshold of 0.5 exhibited the highest correlation with human evaluation.

\begin{table}[htb]
    \centering
    \caption{Correlation between UniDet-based metric with different thresholds and human evaluation on the category of 2D-spatial relationships.}
    \label{tab:unidet_correlation}
    \resizebox{\linewidth}{!}{
    \begin{tabular}{lcccc}
    \toprule
    \textbf{Threshold} & \textbf{0.0}  & \textbf{0.25}  & \textbf{0.50}  & \textbf{0.75} \\ \midrule
    $\tau$ (↑)   & 0.1708 & 0.2060 & \textbf{0.4756} & 0.4634 \\
    $\rho$ (↑)   & 0.2181 & 0.2334 & \textbf{0.5136} & 0.5020 \\
    \bottomrule
    \end{tabular}
    }
\end{table}

\begin{figure}[!htb]
\centering
\includegraphics[width=\linewidth]{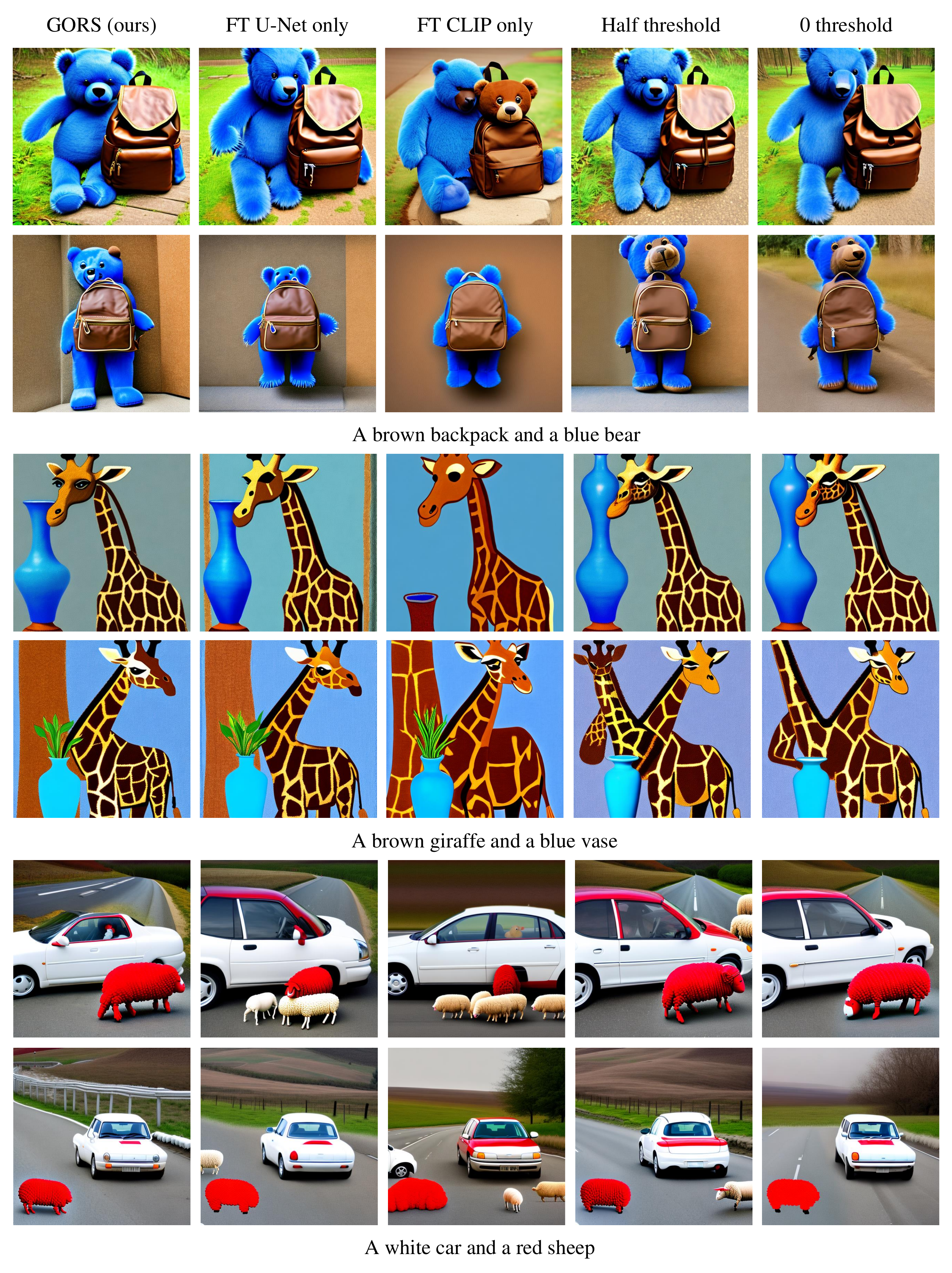}
\caption{Qualitative comparison of ablation study on fine-tuning strategy and threshold.}
\label{fig:ablation}
\end{figure}

\textbf{Qualitative results.}
We show the qualitative results of the variants in ablation study in Figure~\ref{fig:ablation}. When only CLIP is fine-tuned with LoRA, the generated images do not bind attributes to correct objects (for example, Figure~\ref{fig:ablation} Row. 3 Col. 3 and Row. 6 Col. 3). Noticeable improvements are observed in the generated images when U-Net is fine-tuned by LoRA, particularly when both CLIP and U-Net are finetuned together. %
Furthermore, we delve into the effect of the threshold for selecting images aligned with text prompts for fine-tuning. A higher threshold value enables the selection of images that are highly aligned with text prompts for finetuning, ensuring that only well-aligned examples are incorporated into the finetuning process. In contrast, a lower threshold leads to the inclusion of misaligned images during finetuning, which can degrade the compositional ability of the finetuned text-to-image models (for example, Figure~\ref{fig:ablation} last two columns in Row. 2).

\subsection{Limitation and Potential Negative Social Impacts}
\begin{figure*}[ht]
\centering
\includegraphics[width=\linewidth]{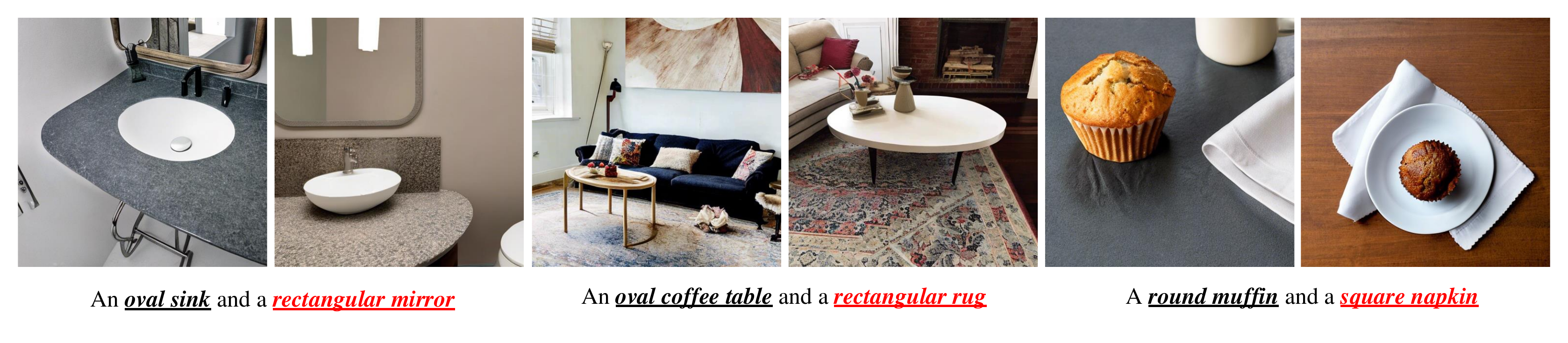}
\caption{Failure cases of the evaluation metric BLIP-VQA.}
\label{fig:failure case}
\end{figure*}

One limitation of our work is the absence of a unified metric for all forms of compositionality. Future research can explore the potential of multimodal LLM to develop a unified metric. 
Our proposed evaluation metrics are not perfect. As shown by the failure cases in Fig.~\ref{fig:failure case}, BLIP-VQA may fail in challenging cases, for example, the objects' shapes are not fully visible in the image, shape's description is uncommon or the objects are not easy to recognize. The UniDet-based evaluation metric is limited to evaluating 2D spatial relationships and we leave 3D spatial relationships for future study.
Researchers need to be aware of the potential negative social impact from the abuse of text-to-image models and the biases of hallucinations from image generators as well as pre-trained multimodal models and multimodal LLMs.
Future research should exercise caution when working with generated images and LLM-generated content and devise appropriate prompts to mitigate the impact of hallucinations and bias in those models.

\section{Conclusion and Discussions}

We propose~\name, a comprehensive benchmark for open-world compositional text-to-image generation, consisting of 8,000 prompts from 4 categories and 8 sub-categories. We propose new evaluation metrics and an improved baseline for the benchmark, and validate the effectiveness of the metrics and method by extensive evaluation. 
We further study the potential and limitations of MLLMs as a unified metrics.
Besides, we propose a simple yet effective way~\abbr~to boost the compositionally of text-to-image models.
When studying generative models, researchers need to be aware of the potential negative social impact, for example, it might be abused to generate fake news. We also need to be aware of the bias from the image generators and the evaluation metrics based on pretrained multimodal models. 

\section*{Acknowledgements}
This work is supported by the National Nature Science Foundation of China (No. 62402406), HKU IDS research Seed Fund, HKU Fintech Academy R\&D Funding, HKU Seed Fund for Basic Research, and HKU Seed Fund for Translational and Applied Research.

%% file: tables/tab_color_shape_test.tex
\begin{table*}[t]
\centering
    \caption{Comparisons of evaluation metrics on attribute binding (color and shape), with scores unnormalized. \textbf{Bold} stands for the best score across 7 models in T2I-Compbench~\cite{huang2024t2i}. \colorbox{mycolor_blue}{Blue} represents the proposed metric for the category, and \colorbox{mycolor_green}{green} stands for the human evaluation, applicable to the following Table~\ref{benchmark:texture_spatial},~\ref{benchmark:3d_spatial_numeracy} and~\ref{benchmark:non_spatial_compre}. 
    } 
\label{benchmark:color_shape}
\resizebox{\linewidth}{!}{%
\begin{tabular}{lccc>{\columncolor{mycolor_blue}}cccc>{\columncolor{mycolor_green}}cccc>{\columncolor{mycolor_blue}}cccc>{\columncolor{mycolor_green}}c}
\toprule
\multicolumn{1}{c}{\multirow{2}{*}{\textbf{Model}}} & \multicolumn{8}{c}{\textbf{Color}}   & \multicolumn{8}{c}{\textbf{Shape}}         \\
              \cmidrule(lr){2-9}\cmidrule(lr){10-17}
\multicolumn{1}{c}{}    & CLIP   & B-CLIP &  B-VQA-n & B-VQA & mGPT-CoT & GPT-4V & Share–CoT  & Human  & CLIP   & B-CLIP & B-VQA-n & B-VQA & mGPT-CoT & GPT-4V & Share–CoT  & Human  \\
              \midrule
Stable v1-4~\cite{rombach2022high}   & 0.3214 & 0.7454 &0.5875  & 0.3765   & 0.7424 & 0.6324 & 0.6185   & 0.6533 & 0.3112 & 0.7077 & 0.6771 & 0.3576   & 0.7197  & 0.4466 & 0.5470  & 0.6160 \\

Stable v2~\cite{rombach2022high}     & 0.3335 & 0.7616 & 0.7249 & 0.5065   & 0.7764  & 0.6717 & 0.7056  & 0.7747 & \textbf{0.3203} & 0.7191 & 0.7517 & 0.4221   & 0.7279  & 0.6212 & 0.5815  & 0.6587 \\

Composable + SD v2~\cite{liu2022compositional}    & 0.3178 & 0.7352 & 0.5744 & 0.4063  & 0.7524 & 0.6295 & 0.6597   & 0.6187 & 0.3092 & 0.6985 & 0.6125 & 0.3299   & 0.7124  & 0.5655 & 0.5339  & 0.5133 \\

Structured + SD v2~\cite{feng2022training}  & 0.3319 & 0.7626 &0.7184 & 0.4990   & 0.7822  & 0.6671 & 0.7253  & 0.7867 & 0.3178 & 0.7177 & 0.7500& 0.4218   & 0.7228   & 0.6348 & 0.5886 & 0.6413 \\

Attn-Exct + SD v2~\cite{chefer2023attend}        & 0.3374 & \textbf{0.7810} & 0.8362 & 0.6400   & \textbf{0.8194}  & \textbf{0.7539} & \textbf{0.8110}  & 0.8240 & 0.3189 & \textbf{0.7209} & 0.7723 & 0.4517   & 0.7299  & 0.5990 & 0.6066  & 0.6360 \\ \midrule

\abbr~ + SD v2 (ours) & \textbf{0.3395} & 0.7681 &\textbf{0.8471} & \textbf{0.6603}   & 0.8067  & 0.7296 & 0.7994  & \textbf{0.8320} & 0.2973 & 0.7201 &\textbf{0.7937} & \textbf{0.4785}   & \textbf{0.7303} & \textbf{0.6563} & \textbf{0.6329}   & \textbf{0.7040}  \\

\bottomrule
\end{tabular}
}
\vspace{0pt}
\end{table*}

%% file: tables/tab_texture_spatial.tex
\begin{table*}[t]
\centering
    \caption{Comparisons of evaluation metrics on attribute binding (texture) and 2D-spatial relationship. %
    } 
\vspace{0pt}
\label{benchmark:texture_spatial}
\resizebox{\linewidth}{!}{%
\begin{tabular}%
{lccc>{\columncolor{mycolor_blue}}cccc>{\columncolor{mycolor_green}}ccc>{\columncolor{mycolor_blue}}cccc>{\columncolor{mycolor_green}}ccc}
\toprule
\multicolumn{1}{c}{\multirow{2}{*}{\textbf{Model}}} & \multicolumn{8}{c}{\textbf{Texture}}                                           & \multicolumn{8}{c}{\textbf{2D-Spatial}}                                           \\
\cmidrule(lr){2-9} \cmidrule(lr){10-16}
\multicolumn{1}{c}{}                                & CLIP   & B-CLIP    & B-VQA-n      & B-VQA        & mGPT-CoT  & GPT-4V & Share–CoT       & Human           & CLIP   & B-CLIP          & UniDet          & mGPT-CoT   & GPT-4V & Share–CoT   & Human           \\
\midrule
Stable v1-4~\cite{rombach2022high}                                         & 0.3081 & 0.7111  & 0.6173        & 0.4156          & 0.7836     & 0.6158 & 0.5728     & 0.7227          & 0.3142 & 0.7667          & 0.1246       & 0.8338     & 0.4297 & 0.7110     & 0.3813          \\
Stable v2~\cite{rombach2022high}                                            & 0.3185 & 0.7240   & 0.7054       & 0.4922          & 0.7851     & 0.6594 & 0.6100     & 0.7827          & 0.3206 & 0.7723          & 0.1342          & 0.8367     & 0.4450 & 0.7197     & 0.3467          \\
Composable + SD v2~\cite{liu2022compositional}                                          & 0.3092 & 0.6995  & 0.5604        & 0.3645          & 0.7588      & 0.5870 & 0.5259    & 0.6333          & 0.3001 & 0.7409          & 0.0800          & 0.8222      & 0.2200 & 0.6633    & 0.3080          \\
Structured + SD v2~\cite{feng2022training}                                        & 0.3167 & 0.7234  & 0.7007        & 0.4900          & 0.7806   & 0.6674 & 0.6108       & 0.7760          & 0.3201 & 0.7726          & 0.1386          & 0.8361     & 0.4617 & 0.7183     & 0.3467          \\
Attn-Exct + SD v2~\cite{chefer2023attend}                                              & 0.3171 & 0.7206   & 0.7830       & 0.5963          & 0.8062        & 0.7077 & 0.6758  & 0.8400          & 0.3213 & 0.7742          & 0.1455          & \textbf{0.8407} & 0.5103 & 0.7330 & 0.4027          \\ \midrule

\abbr~+ SD v2 (ours)                                       & \textbf{0.3233} & \textbf{0.7315} & \textbf{0.7991} & \textbf{0.6287} & \textbf{0.8106} & \textbf{0.7266} & \textbf{0.7054} & \textbf{0.8573} & \textbf{0.3242} & \textbf{0.7854} & \textbf{0.1815} & 0.8362    &  \textbf{0.4640} & \textbf{0.7373}     & \textbf{0.4560}\\

\bottomrule
\end{tabular}
}
\vspace{5pt}
\end{table*}

%% file: tables/tab_num_3d.tex
\begin{table*}[t]
\centering
    \caption{Comparisons of evaluation metrics on 3D-spatial relationship and numeracy task. %
    } 
\vspace{0pt}
\label{benchmark:3d_spatial_numeracy}
\resizebox{\linewidth}{!}{%
\begin{tabular}%
{lcc>{\columncolor{mycolor_blue}}cccc>{\columncolor{mycolor_green}}ccc>{\columncolor{mycolor_blue}}cccc>{\columncolor{mycolor_green}}c}
\toprule
\multicolumn{1}{c}{\multirow{2}{*}{\textbf{Model}}} & \multicolumn{7}{c}{\textbf{3D-Spatial}}                                           & \multicolumn{7}{c}{\textbf{Numeracy}}                                           \\
\cmidrule(lr){2-8} \cmidrule(lr){9-15}
\multicolumn{1}{c}{}                                & CLIP   & B-CLIP    & UniDet  & mGPT-CoT  & GPT-4V & Share–CoT & Human           & CLIP   & B-CLIP    & UniDet          & mGPT-CoT   & GPT-4V & Share–CoT  & Human           \\
\midrule
Stable v1-4~\cite{rombach2022high} & 0.3001 & 0.7344 & 0.3030 & 0.8249 & 0.2880 & 0.6977 & 0.4987 & 0.3021 & 0.6598 & 0.4456 & 0.8427 & 0.3643 & 0.7423 & 0.5200                         \\
Stable v2~\cite{rombach2022high} & 0.3100 & 0.7486 & 0.3230 & 0.8338 & 0.3600 & 0.7160 & 0.5040 & 0.3138 & 0.6588 & 0.4582 & 0.8364 & 0.4320 & 0.7653 & 0.5253                                        \\
Composable + SD v2~\cite{liu2022compositional} & 0.2955 & 0.7300 & 0.2847 & 0.8253 & 0.2413 & 0.6853 & 0.4933 & 0.3028 & 0.6531 & 0.4272 & 0.8391 & 0.3842 & 0.7540 & 0.5187                                        \\
Structured + SD v2~\cite{feng2022training} & 0.3093 & 0.7482 & 0.3224 & 0.8310 & 0.3653 & 0.7153 & 0.5027 & 0.3125 & 0.6588 & 0.4557 & \textbf{0.8440} & 0.4258 & 0.7637 & 0.5680                                        \\
Attn-Exct + SD v2~\cite{chefer2023attend} & 0.3096 & 0.7517 & 0.3222 & \textbf{0.8365} & 0.3633 & 0.7213 & 0.5400 & 0.3130 & 0.6607 & 0.4773 & 0.8439 & 0.4660 & \textbf{0.7857} & 0.5613                                                          \\ \midrule

\abbr~ + SD v2 (ours) & \textbf{0.3140} & \textbf{0.7615} & \textbf{0.3572} & 0.8292 & \textbf{0.3753} & \textbf{0.7327} & \textbf{0.5547} & \textbf{0.3152} & \textbf{0.6715} & \textbf{0.4830} & 0.8376 & \textbf{0.4740} & 0.7847 & \textbf{0.5747}     \\

\bottomrule
\end{tabular}
}
\vspace{0pt}
\end{table*}

%% file: tables/tab_non_spatial_compre.tex
\begin{table*}[t]
\centering
    \caption{Comparisons of evaluation metrics on the non-spatial relationship and complex compositions. %
    } 
    \vspace{0pt}
\label{benchmark:non_spatial_compre}
\resizebox{\linewidth}{!}{ 
\begin{tabular}%
{lccc>{\columncolor{mycolor_blue}}cc>{\columncolor{mycolor_green}}ccccc>{\columncolor{mycolor_blue}}cc>{\columncolor{mycolor_green}}c}
\toprule
\multicolumn{1}{c}{\multirow{2}{*}{\textbf{Model}}} & \multicolumn{6}{c}{\textbf{Non-spatial}}                              & \multicolumn{7}{c}{\textbf{Complex}}                                      \\
              \cmidrule(lr){2-7}\cmidrule(lr){8-14}
\multicolumn{1}{c}{}   & CLIP   & B-CLIP    & mGPT-CoT  & GPT-4V & Share–CoT      & Human           & CLIP   & B-CLIP & 3-in-1 & mGPT-CoT  & GPT-4V & Share–CoT      & Human                                 \\
\midrule
Stable v1-4~\cite{rombach2022high}                                & 0.3079 & 0.7565          & 0.8170       & 0.7717 & 0.7487   & 0.9653          & 0.2876 & 0.6816 & 0.3080           & 0.8075    & 0.6453 & 0.7727      & 0.8067                                \\
Stable v2~\cite{rombach2022high}                                  & 0.3127 & 0.7609          & \textbf{0.8235}       & 0.8153 & 0.7567   & 0.9827          & \textbf{0.3096} & 0.6893 & 0.3386           & 0.8094     & 0.6483 & \textbf{0.7783}     & 0.8480                                \\
Composable + SD v2~\cite{liu2022compositional}                                 & 0.2980 & 0.7038          & 0.7936     & 0.5030 & 0.6927     & 0.8120          & 0.3014 & 0.6638 & 0.2898           & 0.8083   & 0.5637 & 0.7487       & 0.7520                                \\
Structured + SD v2~\cite{feng2022training}                               & 0.3111 & 0.7614          & 0.8221     & 0.8127 & 0.7560     & 0.9773          & 0.3084 & \textbf{0.6902} & 0.3355           & 0.8076   & 0.6400 & 0.7777       & 0.8333                                \\
Attn-Exct + SD v2~\cite{chefer2023attend}                                     & 0.3109 & 0.7607          & 0.8214     & 0.8243 & 0.7593     & 0.9533          & 0.2913 & 0.6875 & \textbf{0.3401}          & 0.8078     & 0.6817 & 0.7763     & 0.8573                       \\ \midrule

\abbr~ + SD v2 (ours)                              & \textbf{0.3193} & \textbf{0.7619}        & 0.8172 & \textbf{0.8420} & \textbf{0.7637} & \textbf{0.9853} & 0.2973 & 0.6841 & 0.3328  & \textbf{0.8095} & \textbf{0.6850} & 0.7737 & \textbf{0.8680}                            \\

\bottomrule
\end{tabular}
}
\vspace{0pt}
\end{table*}

%% file: tables/tab_correlation.tex
\begin{table*}[t]
\centering
    \caption{The correlation between automatic evaluation metrics and human evaluation. Our proposed metrics demonstrate a significant improvement over existing metrics in terms of Kendall's $\tau$ and Spearmanr's $\rho$. \textbf{Bold} stands for the highest correlation score across all non-MLLM metrics. \colorbox{mycolor_red}{Red} indicates the highest correlation score with MLLM metrics included.
    } 

\label{human_corr}
    \vspace{5pt}
\resizebox{\linewidth}{!}{%
\begin{tabular}{l cccc cccc cccc cccc}
\toprule
\multirow{2}{*}{ \textbf{Metric}} & \multicolumn{2}{c}{\textbf{Attribute-color}} & \multicolumn{2}{c}{\textbf{Attribute-shape}} & \multicolumn{2}{c}{\textbf{Attribute-texture}} & \multicolumn{2}{c}
{\textbf{2D Spatial rel}} & \multicolumn{2}{c}
{\textbf{3D Spatial rel}} & \multicolumn{2}{c}
{\textbf{Numeracy}} & \multicolumn{2}{c}
{\textbf{Non-spatial rel}} & \multicolumn{2}{c}
{\textbf{Complex} }\\
\cmidrule(lr){2-3}\cmidrule(lr){4-5}\cmidrule(lr){6-7}\cmidrule(lr){8-9}\cmidrule(lr){10-11}\cmidrule(lr){12-13}
\cmidrule(lr){14-15}\cmidrule(lr){16-17}
 & $\tau(\uparrow)$ & $\rho(\uparrow)$ & $\tau(\uparrow)$ & $\rho(\uparrow)$ & $\tau(\uparrow)$ & $\rho(\uparrow)$& $\tau(\uparrow)$ & $\rho(\uparrow)$ & $\tau(\uparrow)$ & $\rho(\uparrow)$ & $\tau(\uparrow)$ & $\rho(\uparrow)$
 & $\tau(\uparrow)$ & $\rho(\uparrow)$
 & $\tau(\uparrow)$ & $\rho(\uparrow)$\\
    \midrule
    CLIP &0.1938&0.2773&0.0555&0.0821&0.2890&0.4008&0.2741&0.3548
    & 0.2151 & 0.3009 & 0.1557 & 0.2187 
    &\textbf{0.2470}&\textbf{0.3161}&0.0650&0.0847 \\
    B-CLIP  &0.2674&0.3788&0.1692&0.2413&0.2999&0.4187&0.1983&0.2544
    & 0.2042 & 0.2840 & 0.0859 & 0.1257 
    &0.2342&0.2964&0.1963&0.2755 \\
    B-VQA-n &0.4602&0.6179&0.2280&0.3180&0.4227&0.5830&-&-& - & - & - & - 
    &-&-&-&-\\
    B-VQA &\cellcolor{mycolor_red}\textbf{0.6297}&\cellcolor{mycolor_red}\textbf{0.7958}& \textbf{0.2707} & \cellcolor{mycolor_red}\textbf{0.3795} & \cellcolor{mycolor_red}\textbf{0.5177} & \cellcolor{mycolor_red}\textbf{0.6995} & - & -&-&-&-&- & - & - & - & - \\
    UniDet &-&-& - & - & - & - & \cellcolor{mycolor_red}\textbf{0.4756} & \textbf{\cellcolor{mycolor_red}0.5136}&\textbf{\cellcolor{mycolor_red}0.3126}&\textbf{\cellcolor{mycolor_red}0.4262}&\textbf{\cellcolor{mycolor_red}0.4251}&\textbf{\cellcolor{mycolor_red}0.5273}& - & - & - & - \\
    3-in-1 &-&-& - & - & - & - & - & - 
    & - & - & - & -
    &-&-&\textbf{0.2831}&\textbf{0.3853} \\
    \midrule
    mGPT &0.1197& 0.1616 & 0.1282& 0.1775 & 0.1061 & 0.1460 & 0.0208 & 0.0229
    & 0.0980 & 0.1373 & 0.0647 & 0.0967
    &0.1181&0.1418&0.0066&0.0084 \\
    mGPT-CoT &0.3156& 0.4151 & 0.1300& 0.1805 & 0.3453 & 0.4664 & 0.1096 & 0.1239
    & 0.0618 & 0.0760 & 0.1414 & 0.1769
    &0.1944&0.2137&0.1251&0.1463 \\
    Share  & 0.4429 & 0.5260 & 0.1410 & 0.1782 & 0.2858 & 0.3518 & 0.1716  & 0.1837 & 0.1394 & 0.1610 & 0.2908 & 0.3396 & 0.1541 & 0.1801 & 0.0629 & 0.1780 \\
    Share–CoT  & 0.4783 & 0.5847 & \cellcolor{mycolor_red}0.2871 & 0.3659 & 0.4005 & 0.5094 & 0.2889 & 0.3182 & 0.2707 & 0.3186 & 0.4175 & 0.4891 & 0.3401 & 0.3622 & 0.3204 & 0.3649 \\
    GPT-4V & 0.5242 & 0.6465 & 0.2668 & 0.3402 & 0.3944 & 0.4987 & 0.3456 & 0.4038 & 0.2560 & 0.3202 & 0.3777 & 0.4651 & \cellcolor{mycolor_red}0.4756 & \cellcolor{mycolor_red}0.5337 & \cellcolor{mycolor_red}0.5070 & \cellcolor{mycolor_red}0.5942 \\
    \bottomrule
\end{tabular} 
}
    \label{tab:main_statistics_general}
\end{table*}

%% file: tables/tab_new_models.tex
\begin{table*}[t]
\centering
    \caption{Benchmarking on all categories with proposed metrics. \textbf{Bold} stands for the best score across 7 models in T2I-Compbench~\cite{huang2024t2i}. \colorbox{mycolor_red}{Red} indicates the best score across 10 models in~\name. %
    } 
\vspace{0pt}
\label{benchmark:new_model}
\resizebox{\linewidth}{!}{%
\begin{tabular}%
{lccccccccccccccccccc}
\toprule
\multicolumn{1}{c}{\multirow{2}{*}{\textbf{Model}}} & \multicolumn{1}{c}{\textbf{Color}} & \multicolumn{1}{c}{\textbf{Shape}} & \multicolumn{1}{c}{\textbf{Texture}} & \multicolumn{1}{c}{\textbf{2D-Spatial}} & \multicolumn{1}{c}{\textbf{3D-Spatial}} & \multicolumn{1}{c}{\textbf{Numeracy}} & \multicolumn{3}{c}{\textbf{Non-Spatial}} & \multicolumn{3}{c}{\textbf{Complex}}     \\
\cmidrule(lr){2-2} \cmidrule(lr){3-3} \cmidrule(lr){4-4} \cmidrule(lr){5-5} \cmidrule(lr){6-6} \cmidrule(lr){7-7} \cmidrule(lr){8-10} \cmidrule(lr){11-13}
\multicolumn{1}{c}{} & B-VQA   & B-VQA    & B-VQA & UniDet  & UniDet & UniDet & CLIP & GPT-4V & Share-CoT & 3-in-1 & GPT-4V & Share-CoT           \\
\midrule
Stable v1-4~\cite{rombach2022high} & 0.3765 & 0.3576 & 0.4156 & 0.1246 & 0.3030 & 0.4456 & 0.3079 & 0.7717 & 0.7487 & 0.3080 & 0.6453 & 0.7727\\
Stable v2~\cite{rombach2022high}  &  0.5065 & 0.4221 & 0.4922 & 0.1342 & 0.3230 & 0.4582 & 0.3127 & 0.8153 & 0.7567 & 0.3386 & 0.6483 & \textbf{0.7783}\\
Composable + SD v2~\cite{liu2022compositional} & 0.4063 & 0.3299 & 0.3645 & 0.0800 & 0.2847 & 0.4272 & 0.2980 & 0.5030 & 0.6927 & 0.2898 & 0.5637 & 0.7487\\
Structured + SD v2~\cite{feng2022training} &  0.4990 & 0.4218 & 0.4900 & 0.1386 & 0.3224 & 0.4557 & 0.3111 & 0.8127 & 0.7560 & 0.3355 & 0.6400 & 0.7777\\
Attn-Exct + SD v2~\cite{chefer2023attend}  & 0.6400 & 0.4517 & 0.5963 & 0.1455 & 0.3222 & 0.4773 & 0.3109 & 0.8243 & 0.7593 & 0.3401 & 0.6817 & 0.7763\\ \midrule

\abbr-unbiased + SD v2 (ours) & 0.6414 & 0.4546 & 0.6025 & 0.1725 & 0.3300 & \textbf{0.4849} & 0.3158 & \textbf{0.8557} & \textbf{0.7650} & \textbf{0.3470} & 0.6753 & 0.7697\\

\abbr~+ SD v2 (ours) & \textbf{0.6603} & \textbf{0.4785} & \textbf{0.6287} & \textbf{0.1815} & \textbf{0.3572} & 0.4830 & \textbf{0.3193} & 0.8420 & 0.7637 & 0.3328 & \textbf{0.6850} & 0.7737\\
\midrule
Stable XL~\cite{podell2023sdxl} & 0.5879 & 0.4687 & 0.5299 & 0.2133 & 0.3566 & 0.4991 & 0.3119 & 0.8500 & 0.7673 &  0.3237 & 0.7170 & 0.7817\\
Pixart-$\alpha$-ft~\cite{chen2024pixart} & 0.6690 & 0.4927 & 0.6477 & 0.2064 & 0.3901 & 0.5032 & \cellcolor{mycolor_red}0.3197 & 0.8620 & 0.7747 &  0.3433 & 0.7223 & 0.7823\\
DALLE·3~\cite{betker2023improving} & 0.7785 & \cellcolor{mycolor_red}0.6205 & 0.7036 & 0.2865 & 0.3744 & 0.5926 & 0.3003 & 0.9170 & 
\cellcolor{mycolor_red}0.7853 & 
\cellcolor{mycolor_red}0.3773 & \cellcolor{mycolor_red}0.8653 & \cellcolor{mycolor_red}0.7927\\
Stable v3~\cite{esser2024scaling} & \cellcolor{mycolor_red}0.8132 & 0.5885 & \cellcolor{mycolor_red}0.7334 & \cellcolor{mycolor_red}0.3200 & \cellcolor{mycolor_red}0.4084 & 0.6174 & 0.3140 & 0.9093 & 0.7782 &  0.3771 & 0.8717 & 0.7919\\
FLUX.1~\cite{FLUX} & 0.7407 & 0.5718 & 0.6922 & 0.2863 & 0.3866 & \cellcolor{mycolor_red}0.6185 & 0.3127 & \cellcolor{mycolor_red}0.9213 & 0.7809 &  0.3703 & 0.8727 & \cellcolor{mycolor_red}0.7927\\

\bottomrule
\end{tabular}
}
\end{table*}

%% file: tables/tab_minigpt4.tex
\begin{table*}[h]
\centering
    \caption{mGPT benchmarking on 6 sub-categories in \name. %
    } 
\label{app:mGPT}
\resizebox{\linewidth}{!}{%
\begin{tabular}{lcccccccccccc}
\toprule
\multicolumn{1}{c}{\multirow{2}{*}{\textbf{Model}}} & \multicolumn{2}{c}{\textbf{Color}}   & \multicolumn{2}{c}{\textbf{Shape}}    & \multicolumn{2}{c}{\textbf{Texture}}    & \multicolumn{2}{c}{\textbf{Spatial}}  & \multicolumn{2}{c}{\textbf{Non-spatial}} & \multicolumn{2}{c}{\textbf{Complex}} \\
              \cmidrule(lr){2-3}\cmidrule(lr){4-5}\cmidrule(lr){6-7}\cmidrule(lr){8-9}\cmidrule(lr){10-11}\cmidrule(lr){12-13}
\multicolumn{1}{c}{}    &mGPT   &Human &mGPT   &Human &mGPT   &Human &mGPT   &Human &mGPT   &Human &mGPT   &Human  \\
              \midrule  
Stable v1-4~\cite{rombach2022high}        &0.6238  &0.6533  &0.6130  &0.6160  &0.6247  &0.7227  &0.8524  &0.3813  &0.8507  &0.9653  &0.8752  &0.8067  \\
Stable v2~\cite{rombach2022high}          &0.6476  &0.7747  &0.6154  &0.6587  &0.6339  &0.7827  &0.8572  &0.3467  &0.8644  &0.9827  &0.8775  &0.8480  \\
Composable v2~\cite{liu2022compositional} &0.6412  &0.6187  &0.6153  &0.5133  &0.6030  &0.6333  &0.8504  &0.3080  &0.8806  &0.8120  &0.8858  &0.7520  \\
Structured v2~\cite{feng2022training}     &0.6511  &0.7867  &0.6198  &0.6413  &0.6439  &0.7760  &0.8591  &0.3467  &0.8607  &0.9773  &0.8732  &0.8333  \\
Attn-Exct v2~\cite{chefer2023attend}      &\textbf{0.6683}  &0.8240  &0.6175  &0.6360  &0.6482  &0.8400  &0.8536  &0.4027  &0.8684  &0.9533  &0.8725  &0.8573  \\ \midrule

\abbr-unbiased (ours)                           &0.6668  &0.8253  &0.6399  &0.6573  &0.6389  &0.8413  &\textbf{0.8675} &0.4467  &0.8845  &0.9534  &0.8876  &0.8654  \\

\abbr~(ours)                            &0.6677  &\textbf{0.8320}  &\textbf{0.6356}  &\textbf{0.7040}  &\textbf{0.6709}  &\textbf{0.8573}  &0.8584  &\textbf{0.4560}  &\textbf{0.8863}  &\textbf{0.9853}  &\textbf{0.8892}  &\textbf{0.8680}  \\
              \bottomrule
\end{tabular}
}
\end{table*}

%% file: tables/tab_app_seen_unseen.tex
\begin{table}[h]
\centering
\caption{Performances of our model on attribute binding (color, shape, and texture) for seen and unseen sets.}
\label{app:seen_unseen}

\begin{tabular}{@{}lcccccc@{}}
\toprule
\multicolumn{1}{c}{\multirow{2}{*}{\textbf{Metric}}}
         & \multicolumn{2}{c}{\textbf{Color}}         & \multicolumn{2}{c}{\textbf{Shape}}         & \multicolumn{2}{c}{\textbf{Texture}} \\
          \cmidrule(lr){2-3}\cmidrule(lr){4-5}\cmidrule(lr){6-7}
          
         & Seen            & unseen          & Seen            & unseen          & Seen              & unseen  \\
         \midrule
CLIP     & \textbf{0.3422} & 0.3283          & 0.2926          & \textbf{0.3068} & \textbf{0.3240}   & 0.3219  \\
B-CLIP   & \textbf{0.7716} & 0.7612          & \textbf{0.7425} & 0.6752          & \textbf{0.7569}   & 0.6809  \\
B-VQA    & \textbf{0.7192} & 0.5426          & \textbf{0.5500} & 0.3356          & \textbf{0.7647}   & 0.3567  \\
mGPT     & \textbf{0.6626} & 0.6780          & \textbf{0.6381} & 0.6307          & \textbf{0.6773}   & 0.6580  \\
mGPT-CoT & \textbf{0.8082}          & 0.8038 & \textbf{0.7510} & 0.6888          & \textbf{0.8453}   & 0.7412 \\
Share & \textbf{0.8337} & 0.7657 & 0.6014 & \textbf{0.6219} & \textbf{0.6875} & 0.5084 \\
Share-CoT & \textbf{0.7938} & 0.7634 & \textbf{0.6692} & 0.6222 & \textbf{0.7641} & 0.6400 \\
GPT4V & \textbf{0.7491} & 0.6908 & \textbf{0.6824} & 0.6042 & \textbf{0.7916} & 0.5968 \\
\bottomrule
\end{tabular}
\end{table}

%% file: tables/tab_add_training_data.tex
\begin{table}[h]
    \centering
    \caption{Performances of our model on complex compositons on the 3-in-1 metric.}
    \label{tab:add training data}
    \resizebox{\linewidth}{!}{
    
    \begin{tabular}{cccccc}
    
    \toprule
       ours (25) & ours (275) & ours (350)    & ours (700)     &   ours (1050)     &   ours (1400) \\ \midrule
       0.2596 & 0.3086 & 
        0.3299  & 0.3328   &   0.3371   &   \textbf{0.3504} \\
    \bottomrule
    \end{tabular}
   } 
\end{table}

%% file: tables/tab_stability.tex
\begin{table}[t]
\centering
\caption{Stability tests on MLLM metrics.}
\label{stability}
\vspace{5pt}
\small
\begin{tabular}{cccccc}
    \toprule
    \textbf{Test round}    & \textbf{1} & \textbf{2} & \textbf{3}  & \textbf{4} & \textbf{5}     \\
    \midrule
    Share–CoT  & 0.7232 & 0.7109  & 0.7267 & 0.6959 & 0.7129\\
    GPT4V & 0.6918   & 0.7078   & 0.6998 & 0.6758 & 0.6853     \\
    \bottomrule
\end{tabular}
\vspace{5pt}
\end{table}
\vspace{5pt}

%% file: sections/appendix_rebuttal.tex
\newpage
\setcounter{page}{1}
\appendix

\begin{figure}[ht]
\centering
\includegraphics[width=0.75\linewidth]{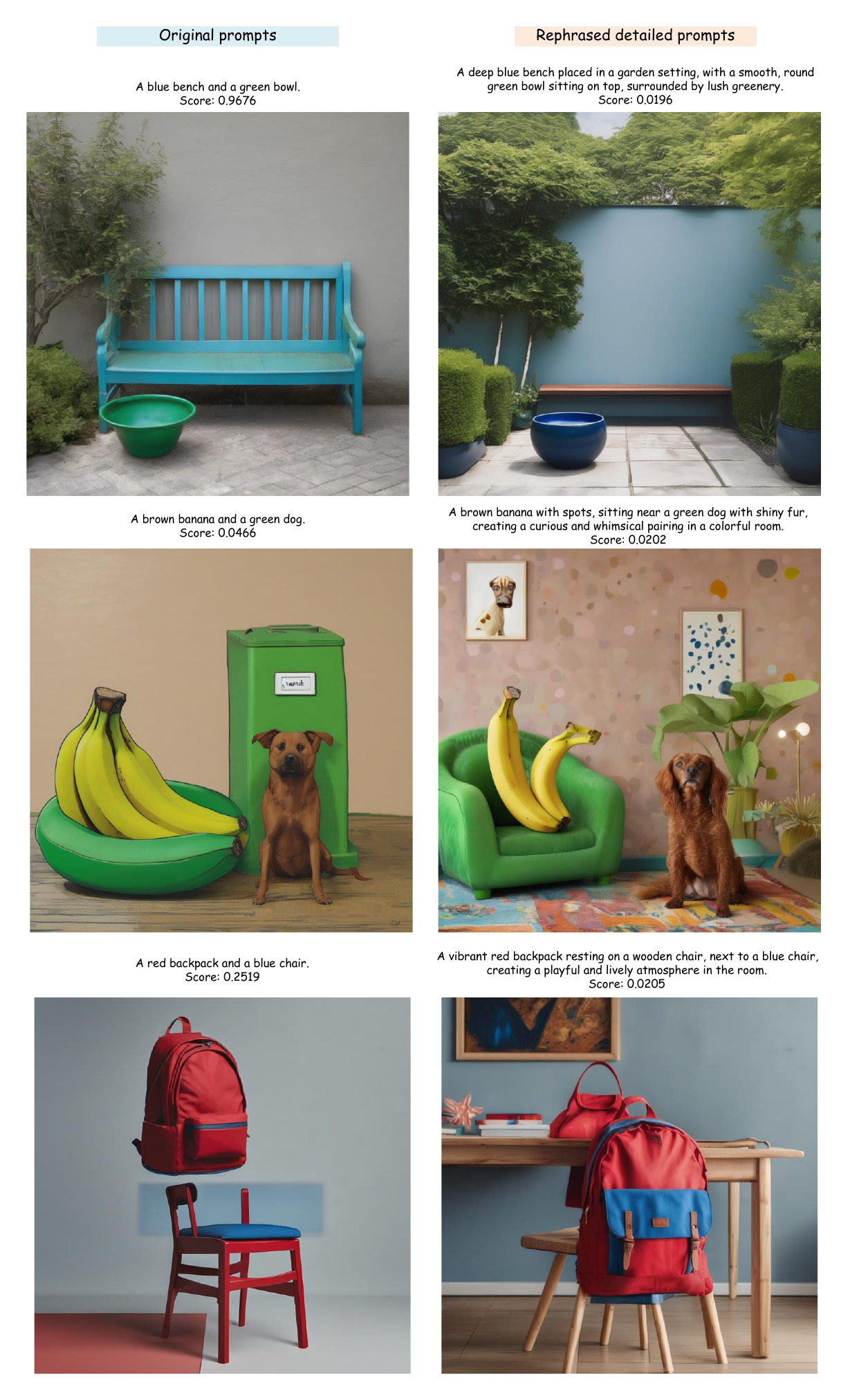}
\caption{Qualitative comparisons between original prompts and rephrased detailed prompts.}
\label{fig:rephrase}
\end{figure}

\textbf{Qualitative comparisons between original prompts and rephrased
detailed prompts.} We observe that enhancing prompt details does not significantly improve the generative outcomes. We use GPT-4 to rephrase the prompts to more detailed versions and compare the evaluation results. The original prompts had an average length of 8.21 words, while the rephrased versions averaged 24.00 words. Qualitative comparisons, shown in Figure~\ref{fig:rephrase}, demonstrate that T2I models continue to face compositional challenges, regardless of the level of detail in the prompts. 

\begin{figure}[ht]
\centering
\includegraphics[width=0.75\linewidth]{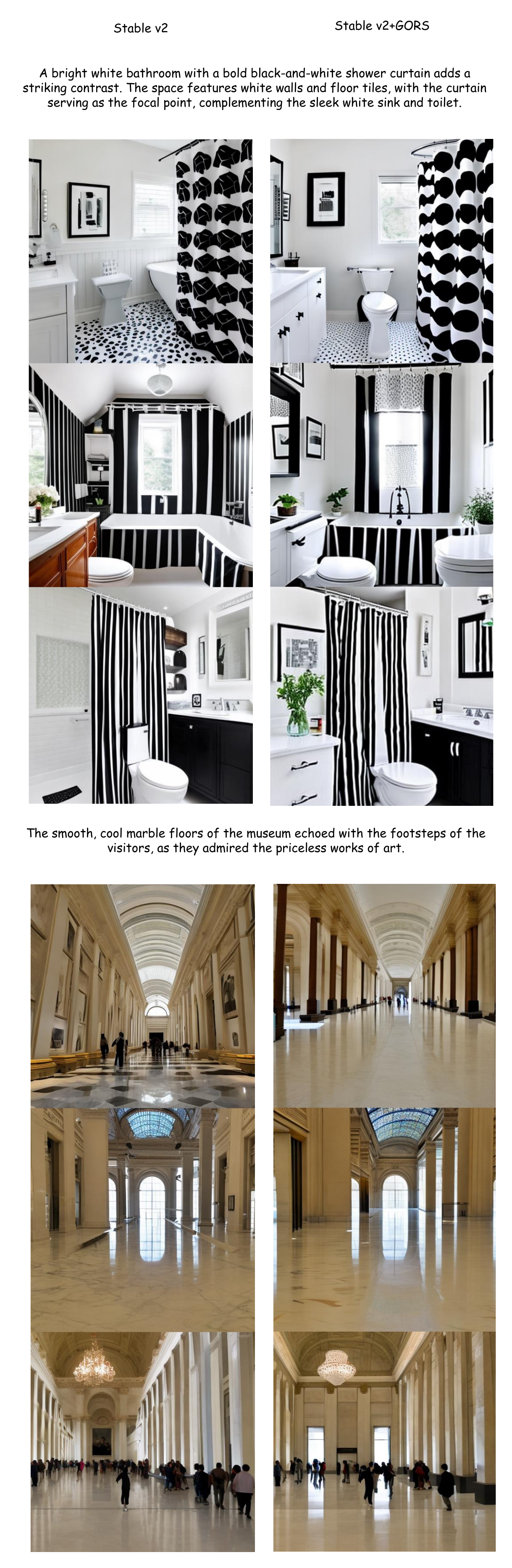}
\caption{Performances of proposed method GORS on short and long text prompts.}
\label{fig:gors_generalization}
\end{figure}

\textbf{Performances of proposed method GORS on short and long text prompts.}
We compare the performances of the base model (Stable v2) and the GORS model (Stable v2+GORS) on long text prompts in Figure~\ref{fig:gors_generalization}. The results show that GORS model is not overly attuned to short prompts and the ability to handle long text prompts is still preserved in our GORS model.

\begin{figure*}[ht]
\centering
\includegraphics[width=0.6\linewidth]{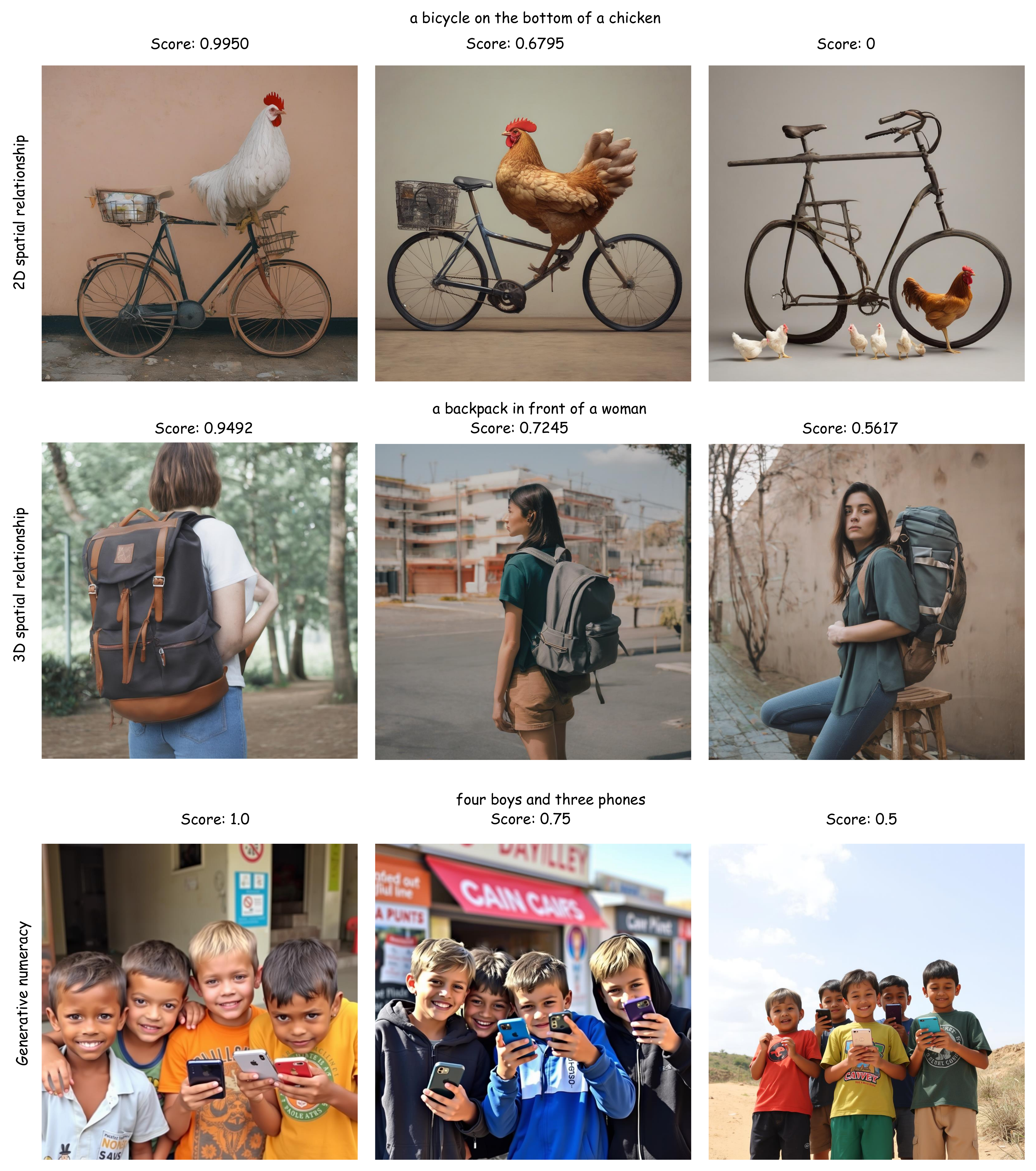}
\caption{Qualitative results of threshold of UniDet-based metrics.}
\label{fig:unidet_threshold}
\end{figure*}

\textbf{Qualitative results of threshold of UniDet-based metrics.} We provide qualitative examples in 2D/3D-spatial relationships and generative numeracy, which use UniDet-based metrics for visualization in Figure~\ref{fig:unidet_threshold}. The results demonstrate that the chosen threshold effectively distinguishes between good and bad cases.

\begin{figure*}[htbp]
\centering
\includegraphics[width=0.9\linewidth]{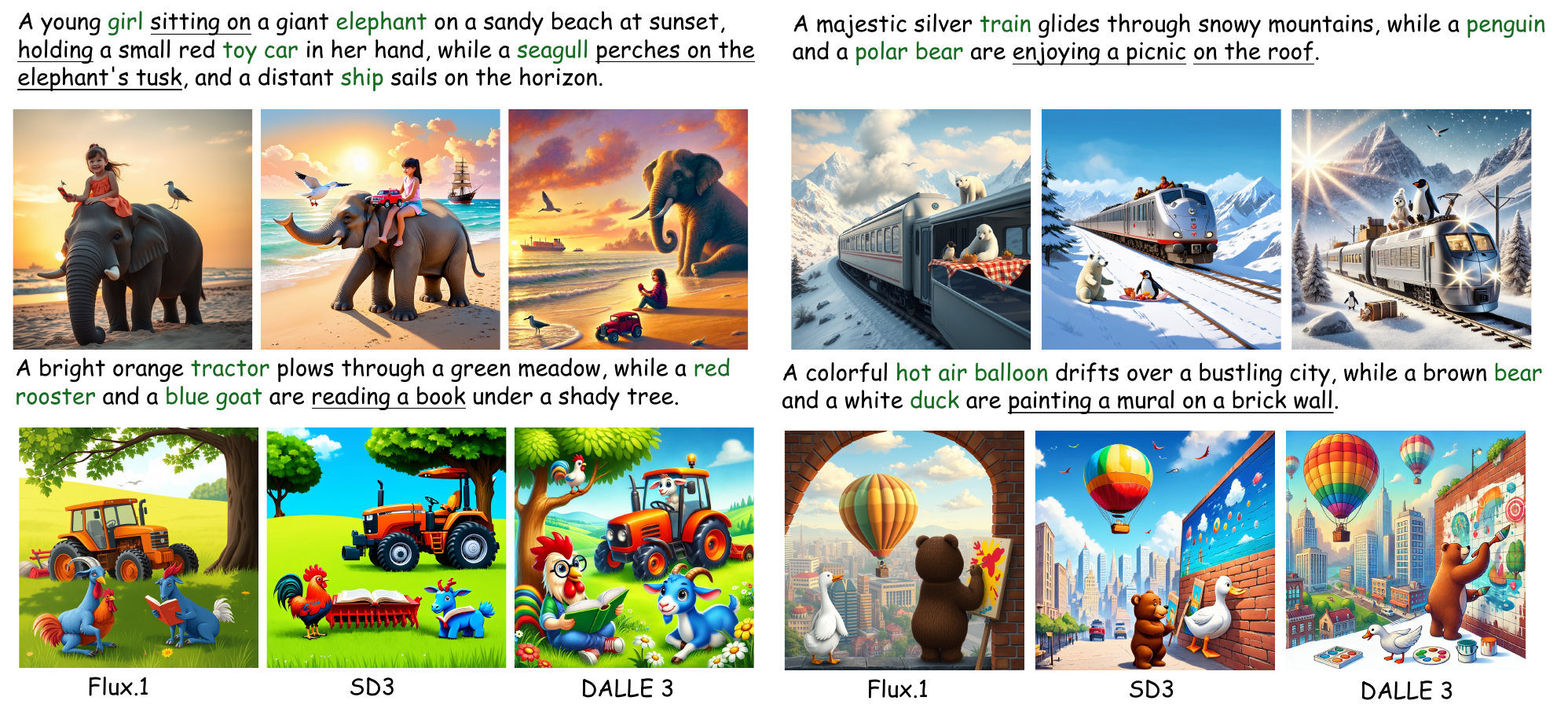}
\caption{Qualitative results of SOTA T2I models (Flux.1, SD3, and DALLE 3) on complex compositional prompts, involving non-spatial relationships and multiple objects (\textgreater=3).}
\label{fig:sd3_dalle3_flux_action_multiobj}
\end{figure*}

\textbf{Qualitative results of SOTA T2I models.} We provide more qualitative examples in non-relationships and multiple objects(\textgreater=3) in Figure~\ref{fig:sd3_dalle3_flux_action_multiobj}. The results demonstrate that even the SOTA models struggle with complex compositional prompts.

%% file: TPAMI_main_rebuttal.bbl
\begin{thebibliography}{10}
\providecommand{\url}[1]{#1}
\csname url@samestyle\endcsname
\providecommand{\newblock}{\relax}
\providecommand{\bibinfo}[2]{#2}
\providecommand{\BIBentrySTDinterwordspacing}{\spaceskip=0pt\relax}
\providecommand{\BIBentryALTinterwordstretchfactor}{4}
\providecommand{\BIBentryALTinterwordspacing}{\spaceskip=\fontdimen2\font plus
\BIBentryALTinterwordstretchfactor\fontdimen3\font minus \fontdimen4\font\relax}
\providecommand{\BIBforeignlanguage}[2]{{%
\expandafter\ifx\csname l@#1\endcsname\relax
\typeout{** WARNING: IEEEtran.bst: No hyphenation pattern has been}%
\typeout{** loaded for the language `#1'. Using the pattern for}%
\typeout{** the default language instead.}%
\else
\language=\csname l@#1\endcsname
\fi
#2}}
\providecommand{\BIBdecl}{\relax}
\BIBdecl

\bibitem{rombach2022high}
R.~Rombach, A.~Blattmann, D.~Lorenz, P.~Esser, and B.~Ommer, ``High-resolution image synthesis with latent diffusion models,'' in \emph{CVPR}, 2022.

\bibitem{ho2022cascaded}
J.~Ho, C.~Saharia, W.~Chan, D.~J. Fleet, M.~Norouzi, and T.~Salimans, ``Cascaded diffusion models for high fidelity image generation,'' \emph{The Journal of Machine Learning Research}, 2022.

\bibitem{saharia2022photorealistic}
C.~Saharia, W.~Chan, S.~Saxena, L.~Li, J.~Whang, E.~L. Denton, K.~Ghasemipour, R.~Gontijo~Lopes, B.~Karagol~Ayan, T.~Salimans \emph{et~al.}, ``Photorealistic text-to-image diffusion models with deep language understanding,'' in \emph{NeurIPS}, 2022.

\bibitem{dhariwal2021diffusion}
P.~Dhariwal and A.~Nichol, ``Diffusion models beat gans on image synthesis,'' in \emph{NeurIPS}, 2021.

\bibitem{nichol2021improved}
A.~Q. Nichol and P.~Dhariwal, ``Improved denoising diffusion probabilistic models,'' \emph{ICLR}, 2021.

\bibitem{chang2023muse}
H.~Chang, H.~Zhang, J.~Barber, A.~Maschinot, J.~Lezama, L.~Jiang, M.-H. Yang, K.~Murphy, W.~T. Freeman, M.~Rubinstein \emph{et~al.}, ``Muse: Text-to-image generation via masked generative transformers,'' \emph{arXiv preprint arXiv:2301.00704}, 2023.

\bibitem{liu2022compositional}
N.~Liu, S.~Li, Y.~Du, A.~Torralba, and J.~B. Tenenbaum, ``Compositional visual generation with composable diffusion models,'' in \emph{ECCV}, 2022.

\bibitem{feng2022training}
W.~Feng, X.~He, T.-J. Fu, V.~Jampani, A.~Akula, P.~Narayana, S.~Basu, X.~E. Wang, and W.~Y. Wang, ``Training-free structured diffusion guidance for compositional text-to-image synthesis,'' in \emph{ICLR}, 2023.

\bibitem{chefer2023attend}
H.~Chefer, Y.~Alaluf, Y.~Vinker, L.~Wolf, and D.~Cohen-Or, ``Attend-and-excite: Attention-based semantic guidance for text-to-image diffusion models,'' in \emph{ACM Trans. Graph.}, 2023.

\bibitem{wu2023harnessing}
Q.~Wu, Y.~Liu, H.~Zhao, T.~Bui, Z.~Lin, Y.~Zhang, and S.~Chang, ``Harnessing the spatial-temporal attention of diffusion models for high-fidelity text-to-image synthesis,'' in \emph{ICCV}, 2023.

\bibitem{radford2021learning}
A.~Radford, J.~W. Kim, C.~Hallacy, A.~Ramesh, G.~Goh, S.~Agarwal, G.~Sastry, A.~Askell, P.~Mishkin, J.~Clark \emph{et~al.}, ``Learning transferable visual models from natural language supervision,'' in \emph{ICML}, 2021.

\bibitem{hessel2021clipscore}
J.~Hessel, A.~Holtzman, M.~Forbes, R.~L. Bras, and Y.~Choi, ``Clipscore: A reference-free evaluation metric for image captioning,'' \emph{arXiv preprint arXiv:2104.08718}, 2021.

\bibitem{li2022blip}
J.~Li, D.~Li, C.~Xiong, and S.~Hoi, ``Blip: Bootstrapping language-image pre-training for unified vision-language understanding and generation,'' in \emph{ICML}, 2022.

\bibitem{li2023blip}
J.~Li, D.~Li, S.~Savarese, and S.~Hoi, ``Blip-2: Bootstrapping language-image pre-training with frozen image encoders and large language models,'' \emph{arXiv preprint arXiv:2301.12597}, 2023.

\bibitem{chen2023sharegpt4v}
L.~Chen, J.~Li, X.~Dong, P.~Zhang, C.~He, J.~Wang, F.~Zhao, and D.~Lin, ``Sharegpt4v: Improving large multi-modal models with better captions,'' \emph{arXiv preprint arXiv:2311.12793}, 2023.

\bibitem{yang2023dawn}
Z.~Yang, L.~Li, K.~Lin, J.~Wang, C.-C. Lin, Z.~Liu, and L.~Wang, ``The dawn of lmms: Preliminary explorations with gpt-4v (ision),'' \emph{arXiv preprint arXiv:2309.17421}, vol.~9, no.~1, p.~1, 2023.

\bibitem{FLUX}
B.~F. Labs, \url{https://github.com/black-forest-labs/flux}, 2024.

\bibitem{esser2024scaling}
P.~Esser, S.~Kulal, A.~Blattmann, R.~Entezari, J.~Müller, H.~Saini, Y.~Levi, D.~Lorenz, A.~Sauer, F.~Boesel, D.~Podell, T.~Dockhorn, Z.~English, K.~Lacey, A.~Goodwin, Y.~Marek, and R.~Rombach, ``Scaling rectified flow transformers for high-resolution image synthesis,'' 2024.

\bibitem{betker2023improving}
J.~Betker, G.~Goh, L.~Jing, T.~Brooks, J.~Wang, L.~Li, L.~Ouyang, J.~Zhuang, J.~Lee, Y.~Guo \emph{et~al.}, ``Improving image generation with better captions,'' \emph{Computer Science. https://cdn. openai. com/papers/dall-e-3. pdf}, vol.~2, no.~3, p.~8, 2023.

\bibitem{chen2024pixart}
J.~Chen, J.~Yu, C.~Ge, L.~Yao, E.~Xie, Y.~Wu, Z.~Wang, J.~Kwok, P.~Luo, H.~Lu \emph{et~al.}, ``Pixart-$alpha$: Fast training of diffusion transformer for photorealistic text-to-image synthesis,'' in \emph{ICLR}, 2024.

\bibitem{podell2023sdxl}
D.~Podell, Z.~English, K.~Lacey, A.~Blattmann, T.~Dockhorn, J.~M{\"u}ller, J.~Penna, and R.~Rombach, ``Sdxl: Improving latent diffusion models for high-resolution image synthesis,'' \emph{arXiv preprint arXiv:2307.01952}, 2023.

\bibitem{huang2024t2i}
K.~Huang, K.~Sun, E.~Xie, Z.~Li, and X.~Liu, ``T2i-compbench: A comprehensive benchmark for open-world compositional text-to-image generation,'' \emph{NeurIPS}, 2024.

\bibitem{bakr2023hrs}
E.~M. Bakr, P.~Sun, X.~Shen, F.~F. Khan, L.~E. Li, and M.~Elhoseiny, ``Hrs-bench: Holistic, reliable and scalable benchmark for text-to-image models,'' in \emph{ICCV}, 2023.

\bibitem{reed2016generative}
S.~Reed, Z.~Akata, X.~Yan, L.~Logeswaran, B.~Schiele, and H.~Lee, ``Generative adversarial text to image synthesis,'' in \emph{ICML}, 2016.

\bibitem{reed2016learning}
S.~E. Reed, Z.~Akata, S.~Mohan, S.~Tenka, B.~Schiele, and H.~Lee, ``Learning what and where to draw,'' in \emph{NeurIPS}, 2016.

\bibitem{zhang2017stackgan}
H.~Zhang, T.~Xu, H.~Li, S.~Zhang, X.~Huang, X.~Wang, and D.~Metaxas, ``Stackgan: Text to photo-realistic image synthesis with stacked generative adversarial networks,'' in \emph{ICCV}, 2017.

\bibitem{xu2018attngan}
T.~Xu, P.~Zhang, Q.~Huang, H.~Zhang, Z.~Gan, X.~Huang, and X.~He, ``Attngan: Fine-grained text to image generation with attentional generative adversarial networks,'' in \emph{CVPR}, 2018.

\bibitem{zhu2019dm}
M.~Zhu, P.~Pan, W.~Chen, and Y.~Yang, ``Dm-gan: Dynamic memory generative adversarial networks for text-to-image synthesis,'' in \emph{CVPR}, 2019.

\bibitem{zhang2021cross}
H.~Zhang, J.~Y. Koh, J.~Baldridge, H.~Lee, and Y.~Yang, ``Cross-modal contrastive learning for text-to-image generation,'' in \emph{CVPR}, 2021.

\bibitem{goodfellow2014generative}
I.~Goodfellow, J.~Pouget-Abadie, M.~Mirza, B.~Xu, D.~Warde-Farley, S.~Ozair, A.~Courville, and Y.~Bengio, ``Generative adversarial nets,'' in \emph{NeurIPS}, 2014.

\bibitem{ramesh2022hierarchical}
A.~Ramesh, P.~Dhariwal, A.~Nichol, C.~Chu, and M.~Chen, ``Hierarchical text-conditional image generation with clip latents,'' \emph{arXiv preprint arXiv:2204.06125}, 2022.

\bibitem{nichol2022glide}
A.~Nichol, P.~Dhariwal, A.~Ramesh, P.~Shyam, P.~Mishkin, B.~McGrew, I.~Sutskever, and M.~Chen, ``Glide: Towards photorealistic image generation and editing with text-guided diffusion models,'' in \emph{ICML}, 2022.

\bibitem{saharia2022imagen}
C.~Saharia, W.~Chan, S.~Saxena, L.~Li, J.~Whang, E.~L. Denton, K.~Ghasemipour, R.~Gontijo~Lopes, B.~Karagol~Ayan, T.~Salimans \emph{et~al.}, ``Photorealistic text-to-image diffusion models with deep language understanding,'' in \emph{NeurIPS}, 2022.

\bibitem{gafni2022make}
O.~Gafni, A.~Polyak, O.~Ashual, S.~Sheynin, D.~Parikh, and Y.~Taigman, ``Make-a-scene: Scene-based text-to-image generation with human priors,'' in \emph{ECCV}, 2022.

\bibitem{gu2023matryoshka}
J.~Gu, S.~Zhai, Y.~Zhang, J.~M. Susskind, and N.~Jaitly, ``Matryoshka diffusion models,'' 2023.

\bibitem{ramesh2021zero}
A.~Ramesh, M.~Pavlov, G.~Goh, S.~Gray, C.~Voss, A.~Radford, M.~Chen, and I.~Sutskever, ``Zero-shot text-to-image generation,'' in \emph{ICML}, 2021.

\bibitem{zhang2023hive}
S.~Zhang, X.~Yang, Y.~Feng, C.~Qin, C.-C. Chen, N.~Yu, Z.~Chen, H.~Wang, S.~Savarese, S.~Ermon \emph{et~al.}, ``Hive: Harnessing human feedback for instructional visual editing,'' \emph{arXiv preprint arXiv:2303.09618}, 2023.

\bibitem{lee2023aligning}
K.~Lee, H.~Liu, M.~Ryu, O.~Watkins, Y.~Du, C.~Boutilier, P.~Abbeel, M.~Ghavamzadeh, and S.~S. Gu, ``Aligning text-to-image models using human feedback,'' \emph{arXiv preprint arXiv:2302.12192}, 2023.

\bibitem{dong2023raft}
H.~Dong, W.~Xiong, D.~Goyal, R.~Pan, S.~Diao, J.~Zhang, K.~Shum, and T.~Zhang, ``Raft: Reward ranked finetuning for generative foundation model alignment,'' \emph{arXiv preprint arXiv:2304.06767}, 2023.

\bibitem{li2022stylet2i}
Z.~Li, M.~R. Min, K.~Li, and C.~Xu, ``Stylet2i: Toward compositional and high-fidelity text-to-image synthesis,'' in \emph{CVPR}, 2022.

\bibitem{patel2023eclipse}
M.~Patel, C.~Kim, S.~Cheng, C.~Baral, and Y.~Yang, ``Eclipse: A resource-efficient text-to-image prior for image generations,'' \emph{arXiv preprint arXiv:2312.04655}, 2023.

\bibitem{liu2024referee}
X.~Liu, T.~Hu, W.~Wang, K.~Kawaguchi, and Y.~Yao, ``Referee can play: An alternative approach to conditional generation via model inversion,'' \emph{arXiv preprint arXiv:2402.16305}, 2024.

\bibitem{park2021benchmark}
D.~H. Park, S.~Azadi, X.~Liu, T.~Darrell, and A.~Rohrbach, ``Benchmark for compositional text-to-image synthesis,'' in \emph{NeurIPS}, 2021.

\bibitem{lian2023llm}
L.~Lian, B.~Li, A.~Yala, and T.~Darrell, ``Llm-grounded diffusion: Enhancing prompt understanding of text-to-image diffusion models with large language models,'' \emph{arXiv preprint arXiv:2305.13655}, 2023.

\bibitem{chen2024training}
M.~Chen, I.~Laina, and A.~Vedaldi, ``Training-free layout control with cross-attention guidance,'' in \emph{WACV}, 2024.

\bibitem{wang2023compositional}
R.~Wang, Z.~Chen, C.~Chen, J.~Ma, H.~Lu, and X.~Lin, ``Compositional text-to-image synthesis with attention map control of diffusion models,'' \emph{arXiv preprint arXiv:2305.13921}, 2023.

\bibitem{meral2023conform}
T.~H.~S. Meral, E.~Simsar, F.~Tombari, and P.~Yanardag, ``Conform: Contrast is all you need for high-fidelity text-to-image diffusion models,'' \emph{arXiv preprint arXiv:2312.06059}, 2023.

\bibitem{kim2023dense}
Y.~Kim, J.~Lee, J.-H. Kim, J.-W. Ha, and J.-Y. Zhu, ``Dense text-to-image generation with attention modulation,'' 2023.

\bibitem{rassin2024linguistic}
R.~Rassin, E.~Hirsch, D.~Glickman, S.~Ravfogel, Y.~Goldberg, and G.~Chechik, ``Linguistic binding in diffusion models: Enhancing attribute correspondence through attention map alignment,'' \emph{NeurIPS}, 2024.

\bibitem{gani2023llm}
H.~Gani, S.~F. Bhat, M.~Naseer, S.~Khan, and P.~Wonka, ``Llm blueprint: Enabling text-to-image generation with complex and detailed prompts,'' \emph{arXiv preprint arXiv:2310.10640}, 2023.

\bibitem{li2023gligen}
Y.~Li, H.~Liu, Q.~Wu, F.~Mu, J.~Yang, J.~Gao, C.~Li, and Y.~J. Lee, ``Gligen: Open-set grounded text-to-image generation,'' in \emph{ICCV}, 2023.

\bibitem{taghipour2024box}
A.~Taghipour, M.~Ghahremani, M.~Bennamoun, A.~M. Rekavandi, H.~Laga, and F.~Boussaid, ``Box it to bind it: Unified layout control and attribute binding in t2i diffusion models,'' \emph{arXiv preprint arXiv:2402.17910}, 2024.

\bibitem{wang2024divide}
Z.~Wang, E.~Xie, A.~Li, Z.~Wang, X.~Liu, and Z.~Li, ``Divide and conquer: Language models can plan and self-correct for compositional text-to-image generation,'' \emph{arXiv preprint arXiv:2401.15688}, 2024.

\bibitem{chen2023reason}
X.~Chen, Y.~Liu, Y.~Yang, J.~Yuan, Q.~You, L.-P. Liu, and H.~Yang, ``Reason out your layout: Evoking the layout master from large language models for text-to-image synthesis,'' \emph{arXiv preprint arXiv:2311.17126}, 2023.

\bibitem{yang2024mastering}
L.~Yang, Z.~Yu, C.~Meng, M.~Xu, S.~Ermon, and B.~Cui, ``Mastering text-to-image diffusion: Recaptioning, planning, and generating with multimodal llms,'' \emph{arXiv preprint arXiv:2401.11708}, 2024.

\bibitem{zhang2024realcompo}
X.~Zhang, L.~Yang, Y.~Cai, Z.~Yu, J.~Xie, Y.~Tian, M.~Xu, Y.~Tang, Y.~Yang, and B.~Cui, ``Realcompo: Dynamic equilibrium between realism and compositionality improves text-to-image diffusion models,'' \emph{arXiv preprint arXiv:2402.12908}, 2024.

\bibitem{wah2011caltech}
C.~Wah, S.~Branson, P.~Welinder, P.~Perona, and S.~Belongie, ``The caltech-ucsd birds-200-2011 dataset,'' 2011.

\bibitem{nilsback2008automated}
M.-E. Nilsback and A.~Zisserman, ``Automated flower classification over a large number of classes,'' in \emph{ICVGIP}, 2008.

\bibitem{lin2014microsoft}
T.-Y. Lin, M.~Maire, S.~Belongie, J.~Hays, P.~Perona, D.~Ramanan, P.~Doll{\'a}r, and C.~L. Zitnick, ``Microsoft coco: Common objects in context,'' in \emph{ECCV}, 2014.

\bibitem{cho2023dall}
J.~Cho, A.~Zala, and M.~Bansal, ``Dall-eval: Probing the reasoning skills and social biases of text-to-image generation models,'' in \emph{ICCV}, 2023.

\bibitem{petsiuk2022human}
V.~Petsiuk, A.~E. Siemenn, S.~Surbehera, Z.~Chin, K.~Tyser, G.~Hunter, A.~Raghavan, Y.~Hicke, B.~A. Plummer, O.~Kerret \emph{et~al.}, ``Human evaluation of text-to-image models on a multi-task benchmark,'' \emph{arXiv preprint arXiv:2211.12112}, 2022.

\bibitem{salimans2016improved}
T.~Salimans, I.~Goodfellow, W.~Zaremba, V.~Cheung, A.~Radford, and X.~Chen, ``Improved techniques for training gans,'' in \emph{NeurIPS}, 2016.

\bibitem{heusel2017gans}
M.~Heusel, H.~Ramsauer, T.~Unterthiner, B.~Nessler, and S.~Hochreiter, ``Gans trained by a two time-scale update rule converge to a local nash equilibrium,'' in \emph{NeurIPS}, 2017.

\bibitem{lu2023llmscore}
Y.~Lu, X.~Yang, X.~Li, X.~E. Wang, and W.~Y. Wang, ``Llmscore: Unveiling the power of large language models in text-to-image synthesis evaluation,'' \emph{arXiv preprint arXiv:2305.11116}, 2023.

\bibitem{chen2023xiqe}
Y.~Chen, ``X-iqe: explainable image quality evaluation for text-to-image generation with visual large language models,'' \emph{arXiv preprint arXiv:2305.10843}, 2023.

\bibitem{wen2023improving}
S.~Wen, G.~Fang, R.~Zhang, P.~Gao, H.~Dong, and D.~Metaxas, ``Improving compositional text-to-image generation with large vision-language models,'' \emph{arXiv preprint arXiv:2310.06311}, 2023.

\bibitem{xu2023imagereward}
J.~Xu, X.~Liu, Y.~Wu, Y.~Tong, Q.~Li, M.~Ding, J.~Tang, and Y.~Dong, ``Imagereward: Learning and evaluating human preferences for text-to-image generation,'' 2023.

\bibitem{sun2023dreamsync}
J.~Sun, D.~Fu, Y.~Hu, S.~Wang, R.~Rassin, D.-C. Juan, D.~Alon, C.~Herrmann, S.~van Steenkiste, R.~Krishna \emph{et~al.}, ``Dreamsync: Aligning text-to-image generation with image understanding feedback,'' \emph{arXiv preprint arXiv:2311.17946}, 2023.

\bibitem{kirstain2024pick}
Y.~Kirstain, A.~Polyak, U.~Singer, S.~Matiana, J.~Penna, and O.~Levy, ``Pick-a-pic: An open dataset of user preferences for text-to-image generation,'' \emph{NeurIPS}, 2024.

\bibitem{wu2023better}
X.~Wu, K.~Sun, F.~Zhu, R.~Zhao, and H.~Li, ``Better aligning text-to-image models with human preference,'' 2023.

\bibitem{wu2023human}
X.~Wu, Y.~Hao, K.~Sun, Y.~Chen, F.~Zhu, R.~Zhao, and H.~Li, ``Human preference score v2: A solid benchmark for evaluating human preferences of text-to-image synthesis,'' \emph{arXiv preprint arXiv:2306.09341}, 2023.

\bibitem{liang2023rich}
Y.~Liang, J.~He, G.~Li, P.~Li, A.~Klimovskiy, N.~Carolan, J.~Sun, J.~Pont-Tuset, S.~Young, F.~Yang \emph{et~al.}, ``Rich human feedback for text-to-image generation,'' \emph{arXiv preprint arXiv:2312.10240}, 2023.

\bibitem{ku2023imagenhub}
M.~Ku, T.~Li, K.~Zhang, Y.~Lu, X.~Fu, W.~Zhuang, and W.~Chen, ``Imagenhub: Standardizing the evaluation of conditional image generation models,'' \emph{arXiv preprint arXiv:2310.01596}, 2023.

\bibitem{ku2023viescore}
M.~Ku, D.~Jiang, C.~Wei, X.~Yue, and W.~Chen, ``Viescore: Towards explainable metrics for conditional image synthesis evaluation,'' \emph{arXiv preprint arXiv:2312.14867}, 2023.

\bibitem{lee2024holistic}
T.~Lee, M.~Yasunaga, C.~Meng, Y.~Mai, J.~S. Park, A.~Gupta, Y.~Zhang, D.~Narayanan, H.~Teufel, M.~Bellagente \emph{et~al.}, ``Holistic evaluation of text-to-image models,'' \emph{NeurIPS}, 2024.

\bibitem{ChatGPT}
OpenAI, \url{https://openai.com/blog/chatgpt/}, 2023.

\bibitem{zhu2023minigpt4}
D.~Zhu, J.~Chen, X.~Shen, X.~Li, and M.~Elhoseiny, ``Minigpt-4: Enhancing vision-language understanding with advanced large language models,'' \emph{arXiv preprint arXiv:2304.10592}, 2023.

\bibitem{wei2022chain}
J.~Wei, X.~Wang, D.~Schuurmans, M.~Bosma, F.~Xia, E.~Chi, Q.~V. Le, D.~Zhou \emph{et~al.}, ``Chain-of-thought prompting elicits reasoning in large language models,'' in \emph{NeurIPS}, 2022.

\bibitem{zhou2022simple}
X.~Zhou, V.~Koltun, and P.~Kr{\"a}henb{\"u}hl, ``Simple multi-dataset detection,'' in \emph{CVPR}, 2022.

\bibitem{ranftl2021vision}
R.~Ranftl, A.~Bochkovskiy, and V.~Koltun, ``Vision transformers for dense prediction,'' in \emph{ICCV}, 2021.

\bibitem{achiam2023gpt}
J.~Achiam, S.~Adler, S.~Agarwal, L.~Ahmad, I.~Akkaya, F.~L. Aleman, D.~Almeida, J.~Altenschmidt, S.~Altman, S.~Anadkat \emph{et~al.}, ``Gpt-4 technical report,'' \emph{arXiv preprint arXiv:2303.08774}, 2023.

\bibitem{hu2021lora}
E.~J. Hu, Y.~Shen, P.~Wallis, Z.~Allen-Zhu, Y.~Li, S.~Wang, L.~Wang, and W.~Chen, ``Lora: Low-rank adaptation of large language models,'' \emph{arXiv preprint arXiv:2106.09685}, 2021.

\bibitem{teney2020value}
D.~Teney, E.~Abbasnejad, K.~Kafle, R.~Shrestha, C.~Kanan, and A.~Van Den~Hengel, ``On the value of out-of-distribution testing: An example of goodhart's law,'' \emph{Advances in neural information processing systems}, vol.~33, pp. 407--417, 2020.

\bibitem{liu2023grounding}
S.~Liu, Z.~Zeng, T.~Ren, F.~Li, H.~Zhang, J.~Yang, C.~Li, J.~Yang, H.~Su, J.~Zhu, and L.~Zhang, ``Grounding dino: Marrying dino with grounded pre-training for open-set object detection,'' \emph{arXiv preprint arXiv:2303.05499}, 2023.

\bibitem{li2022grounded}
L.~H. Li, P.~Zhang, H.~Zhang, J.~Yang, C.~Li, Y.~Zhong, L.~Wang, L.~Yuan, L.~Zhang, J.-N. Hwang \emph{et~al.}, ``Grounded language-image pre-training,'' in \emph{CVPR}, 2022.

\bibitem{shao2019objects365}
S.~Shao, Z.~Li, T.~Zhang, C.~Peng, G.~Yu, X.~Zhang, J.~Li, and J.~Sun, ``Objects365: A large-scale, high-quality dataset for object detection,'' in \emph{ICCV}, 2019.

\bibitem{kuznetsova2020open}
A.~Kuznetsova, H.~Rom, N.~Alldrin, J.~Uijlings, I.~Krasin, J.~Pont-Tuset, S.~Kamali, S.~Popov, M.~Malloci, A.~Kolesnikov \emph{et~al.}, ``The open images dataset v4: Unified image classification, object detection, and visual relationship detection at scale,'' in \emph{IJCV}, 2020.

\bibitem{neuhold2017mapillary}
G.~Neuhold, T.~Ollmann, S.~Rota~Bulo, and P.~Kontschieder, ``The mapillary vistas dataset for semantic understanding of street scenes,'' in \emph{ICCV}, 2017.

\bibitem{TP-toolbox-web}
\url{https://github.com/huggingface/diffusers/blob/main/examples/text_to_image}.

\bibitem{loshchilov2019decoupled}
I.~Loshchilov and F.~Hutter, ``Decoupled weight decay regularization,'' in \emph{ICLR}, 2018.

\end{thebibliography}
